\documentclass[sigconf]{aamas} 

\usepackage{balance} 
\usepackage{subcaption}
\usepackage[ruled,vlined]{algorithm2e}
\usepackage{algorithmic}

\setcopyright{none}
\renewcommand\footnotetextcopyrightpermission[1]{}

\acmSubmissionID{1093}

\title[AAMAS-2026]{Extending Multi-Source Bayesian Optimization With Causality Principles}

\author{Luuk Jacobs}
\affiliation{
  \institution{Radboud University}
  \city{Nijmegen}
  \country{The Netherlands}}
\email{luuk.jacobs@ru.nl}

\author{Mohammad Ali Javidian}
\affiliation{
  \institution{Appalachian State University}
  \city{Boone}
  \country{United States}}
\email{javidianma@appstate.edu}
\thanks{An extended abstract version of this work was accepted for the
\textit{Proceedings of the 25th International Conference on Autonomous Agents
and Multiagent Systems (AAMAS 2026)}.}

\begin{abstract}

Multi-Source Bayesian Optimization (MSBO) serves as a variant of the traditional Bayesian Optimization (BO) framework applicable to situations involving optimization of an objective black-box function over multiple information sources such as simulations, surrogate models, or real-world experiments. However, traditional MSBO assumes the input variables of the objective function to be independent and identically distributed, limiting its effectiveness in scenarios where causal information is available and interventions can be performed, such as clinical trials or policy-making. In the single-source domain, Causal Bayesian Optimization (CBO) extends standard BO with the principles of causality, enabling better modeling of variable dependencies. This leads to more accurate optimization, improved decision making, and more efficient use of low-cost information sources. In this article, we propose a principled integration of the MSBO and CBO methodologies in the multi-source domain, leveraging the strengths of both to enhance optimization efficiency and reduce computational complexity in higher-dimensional problems. We present the theoretical foundations of both Causal and Multi-Source Bayesian Optimization, and demonstrate how their synergy informs our Multi-Source Causal Bayesian Optimization (MSCBO) algorithm. We compare the performance of MSCBO against its foundational counterparts for both synthetic and real-world datasets with varying levels of noise, highlighting the robustness and applicability of MSCBO. Based on our findings, we conclude that integrating MSBO with the causality principles of CBO facilitates dimensionality reduction and lowers operational costs, ultimately improving convergence speed, performance, and scalability.

\end{abstract}

\keywords{Bayesian Optimization, Causality, Interventions, Multi-source information, Gaussian Process, Structural equation model, DAG}

\newcommand{\BibTeX}{\rm B\kern-.05em{\sc i\kern-.025em b}\kern-.08em\TeX}

\begin{document}


\pagestyle{fancy}
\fancyhead{}

\maketitle 


\section{Introduction}\label{sec:intro}

As a result of an increase in both the variety of measurement tools and their availability, coupled with a better understanding of problem formulation, modern optimization problems such as clinical trials, manufacturing processes or hyperparameter tuning in machine learning models regularly involve higher variable dimensionalities. As the number of variables increases, so does the search space for the best value assignment to these variables. This exponential growth renders exhaustive experimentation or simulation in search of an optimal configuration expensive and, in most cases, intractable.

A common strategy to reduce this complexity revolves around selectively manipulating certain variables and observing the resulting outcomes. This approach, often referred to as intervention, allows researchers to focus on a smaller and more relevant subset of variables, improving both efficiency and interpretability \citep{roberts2024causalbo}. Among others, clinical trials and research efforts use these interventions \citep{friedman2015fundamentals,steckler2002process}. For instance, in clinical trials, researchers typically administer a new drug to a test group while giving a placebo to a control group. By observing the difference in outcomes between groups, the effect of the drug can be isolated without testing all possible treatment combinations. To make such interventions more systematic, it helps to understand the causal relationships between the variables of the optimization problem. To illustrate this concept, we provide a toy example. Assume that in a factory producing chocolate bars, we know that the temperature of the chocolate during mixing directly affects the final texture. Furthermore, the temperature inside the factory influences the chocolate temperature, thus indirectly affecting its texture as well. If we were to manipulate the temperature of the chocolate directly by installing a temperature controller on the mixer, we would no longer have to worry about the influence of factory temperature, effectively reducing the dimensionality of our problem. Thus, by leveraging the available causal information, we can perform informed interventions that significantly reduce the size of our problem. However, despite their usefulness, interventions are typically resource-intensive, and the challenge becomes even greater in high-dimensional problems where many combinations of interventions may need to be explored.

\subsection{Applying Bayesian Optimization}

Bayesian optimization (BO) has emerged as a powerful framework to address this challenge \citep{mockus1974bayesian}. By modeling the objective function (for example, the efficacy of a medicine or texture of the chocolate) as a probabilistic surrogate, such as a Gaussian process (GP), BO enables the prediction of outcomes and guides the selection of the most informative interventions. This process balances exploration (searching uncertain regions of the input space) and exploitation (focusing on promising areas) to optimize the objective function with minimal evaluations. Its success has made BO a widely used tool in applications where evaluations are expensive.

Despite its widespread adoption, traditional BO assumes independent and identically distributed relationships between variables, which limits its applicability to problems with inherent causal dependencies. In many real-world scenarios, the underlying system is governed by a causal structure that determines how interventions on certain variables cascade to influence others. Ignoring these causal relationships can result in suboptimal solutions and unnecessary interventions \citep{aglietti2020causal}. To address this limitation, recent advances have introduced causal Bayesian optimization (CBO) \citep{aglietti2020causal}, a framework that incorporates causal information into the optimization process. Through the use of this causal knowledge, CBO identifies the minimal set of variables on which to intervene, often a subset of all intervenable variables, and prioritizes interventions based on their causal impact.

In addition, real-world optimization problems frequently involve multiple sources of information with varying levels of fidelity, cost, and reliability. For example, in clinical trials, information can be obtained from preclinical studies, small-scale human trials, or large-scale population data. Integrating these heterogeneous sources effectively requires extending the Bayesian optimization framework to a multi-source setting. Multi-source Bayesian Optimization (MSBO) \citep{frazier2018tutorial, poloczek2017multi}, combines this information from multiple sources to efficiently optimize the objective function, balancing the trade-offs between cost and accuracy.

However, despite these recent advances, BO remains a fairly slow and dimensionality-sensitive approach. In high-dimensional problem settings, traditional BO methods have the tendency to produce prohibitive computation times, rendering convergence infeasible \citep{li2017dropout, shen2023highdim}. Thus, the field remains open for new approaches and methodologies that might speed up the optimization process. In single-source situations, CBO presents itself as a causality-based alternative. However, in the multi-source domain, no such implementation exists. Therefore, the research question we propose is the following. Does integrating causal knowledge from multiple information sources improve the convergence rate and performance of classical Bayesian Optimization in multi-source domains?

\subsection{The integration of CBO and MSBO}\label{sec:integration}

\begin{figure}[!htb]
\centering
\includegraphics[scale=0.5]{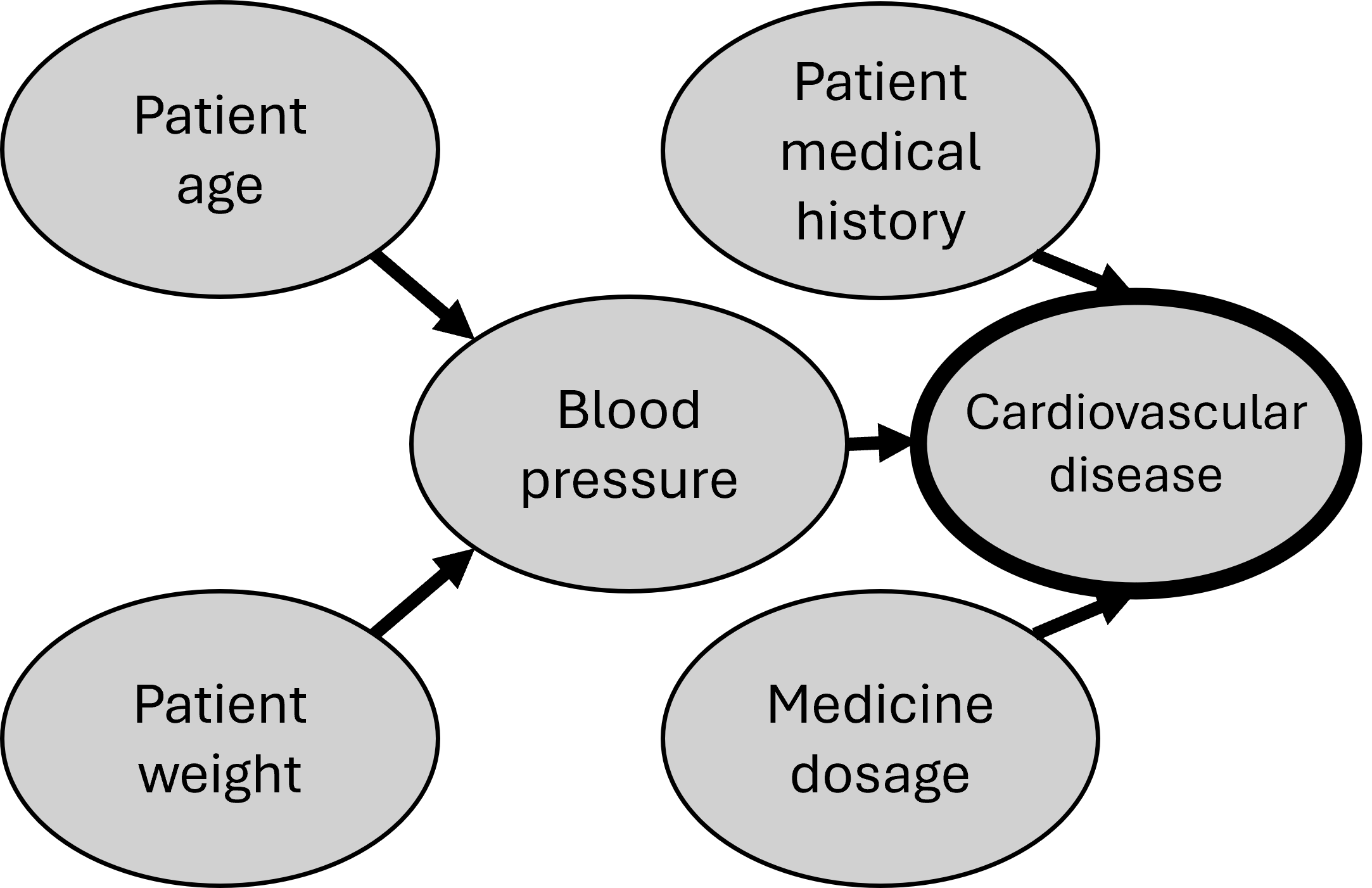}
\caption{Example DAG for cardiovascular disease treatment. Gray nodes represent variables that can be intervened upon. The output variable Cardiovascular Disease is denoted with a thick-dashed node.}
\label{fig:CVD}
\end{figure}

Our research operates on the premise of creating synergy between CBO and MSBO through principled integration. We provide a motivational example that demonstrates how classical MSBO, or a naive integration of CBO, fails to achieve the optimal solution without exhausting the available budget for collecting observational and interventional data:

Assume the situation in which we propose a novel medicine for common cardiovascular diseases. In addition, assume that previous cardiovascular research has provided us with a general understanding of what causal factors influence the emergence of these diseases and how our treatment could play a role in this process. Ideally, to determine our ideal medicine dosage and improve its concepts, one would conduct a clinical intervention study among patients affected by these diseases. A strongly simplified overview of the factors that could influence disease development can be found in Figure \ref{fig:CVD}. We can find patient groups affected by cardiovascular diseases in specialized hospitals, thus introducing multi-source possibilities to the optimization process. However, there may be differences between these hospitals. For example, Hospital A might use different measuring devices or medicine prescription policies compared to Hospital B, leading to datasets with varying levels of usefulness and associated collection costs. Our goal is to gather the maximum possible amount of information within a predefined budget. Performing an intervention within one of these hospitals costs us a total of \$30.000 for each of the variables in which we want to intervene. For example, if we want to test our treatment in a group that has a history of cardiovascular problems, we pay \$30.000 in costs for data collection, filtering and recruiting of the test group.

\begin{figure}[!htb]
\centering
\includegraphics[scale=0.55]{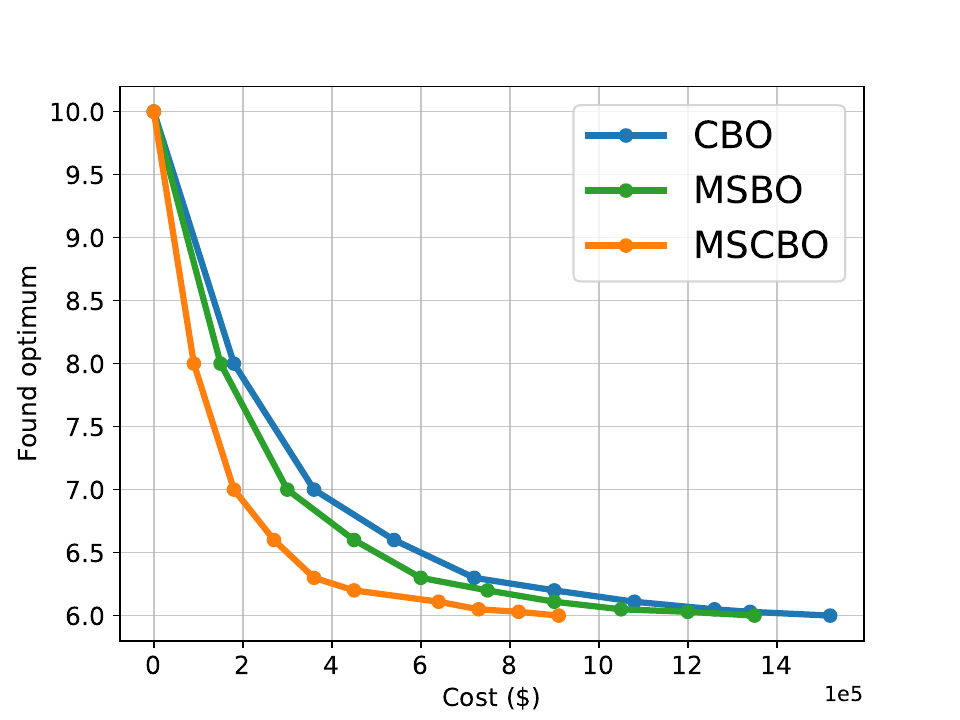}
\caption{A visualization of the theoretical comparison between each of the optimization methods as provided in Section \ref{sec:integration}. We assume the found optimum over each iteration to be identical for the three algorithms.}
\label{fig:TCOMP}
\end{figure}

If we were to look at the problem from a naive point of view, we may decide to simply apply our CBO implementation to each of the sources to determine what our best possible intervention would be. However, this application of CBO to all sources falls short in that it fails to encapsulate inter-source dependencies, differences in operation costs, and most importantly, the capability of switching the search effort for an ideal treatment between sources. Such an algorithm will rapidly exhaust the budget, because it continues the intervention process in each of the hospitals, regardless of how they are contributing to our optimization process as a whole. CBO will identify that, based on the estimation of causal effects, intervening on the observed variables \{blood pressure, treatment, patient medical history\} nullifies the influence of the other two \{age, weight\}. Because we intervene in both sources, a single optimization step will cost us \((\$30.000\ \cdot\ 3\ \text{variables}\ \cdot\ 2\ \text{sources}\ =\ \textbf{\$180.000})\) per intervention. Alternatively, we could use MSBO as is. The problem that arises in this case is that we disregard the causal relationships underlying our disease pattern and thus cannot reduce our intervention sets. As a result, a single optimization step will cost us \((\$30.000\ \cdot\ 5\ \text{variables}\ \cdot\ 1\ \text{source}\ =\ \textbf{\$150.000})\) per intervention. Thus, both a naive integration of CBO and standard MSBO fail to harness the potential synergy capabilities of the two methods, leading to incorrect prioritization of interventions and inefficient use of data sources. In contrast, our principled integration combines causal exploration set reduction based on effect estimation with the source selection mechanism of the two methods. Thus, while information from all sources contributes to the model update, interventions are performed only within the minimal intervention set of the cost-sensitive best source, resulting in an intervention cost of \((\$30.000\ \cdot\ 3\ \text{variables}\ \cdot\ 1\ \text{source}\ =\ \textbf{\$90.000})\) per intervention. Assuming that all three algorithms reach the same optima, this means that an integration of MSBO and CBO operates at a significantly better cost efficiency than its foundational counterparts in a multi-source scenario. An example of what the optimization process would look like for each of the three methods, given that their found optima are identical for each iteration, is provided in Figure \ref{fig:TCOMP}. The figure illustrates how MSCBO produces better optima at lower cost rates in comparison to its predecessors. For example, if the hospital was to allocate approximately \$400.000 in funds to cardiovascular research, it is expected that MSCBO will produce a better optimum within these budget constraints, as can be observed from the figure. 

In this paper, we explore the integration of CBO and MSBO to address the challenges posed by costly, high-dimensional interventions in complex systems. By leveraging causal structures and multi-source information, we aim to develop an optimization framework that minimizes resource usage while achieving superior performance.

\subsubsection{Related work}
Our research is embedded in the general domain of problem-solving with the use of BO techniques, which has risen to prominence from the mid-1970's onward \citep{mockus1974bayesian,mockus1989bayesian}. Recent advancements focused on higher-dimensional problems are often intertwined with novel AI concepts, such as hyperparameter learning for Machine Learning (ML) models \citep{ranjit2019efficient,turner2021bayesian}. In 2018, \citet{frazier2018tutorial} published "A tutorial on Bayesian optimization", detailing the process of designing a BO-algorithm and extending its usage to multi-fidelity and multi-source situations. Such variations have been used in the calibration of building simulation models \citep{zhan2022calibrating}, and
have been benchmarked on hyperparameter tuning and the Rosenbrock function \citep{poloczek2017multi}. The integration of causality with BO-related decision-making systems is fairly novel and includes applications to Multi-armed Bandit (MAB) problems \citep{bareinboim2015bandits} and reinforcement learning \citep{buesing2018woulda}. CBO as a conceptual framework was introduced by \citet{aglietti2020causal} and later translated to a Python package \citep{roberts2024causalbo}. Multi-source CBO operates in settings where information from structurally similar but possibly noisy sources is leveraged to optimize an unknown objective function. This naturally aligns with the core motivations of transfer learning and domain adaptation, where the goal is to improve performance in a target domain by transferring knowledge from auxiliary domains with varying distributions or data characteristics. Recent work in causal domain adaptation highlights how leveraging causal structures enables principled adjustments for distribution shifts across domains \citep{chen2021domain}. This resonates with the framework of our research, in which each source can be seen as a distinct domain arising from modified mechanisms or partial observability, and causal models help guide meaningful transfer across these domains.

\subsubsection{Our Contributions}
In this article, we make the following contributions.  
\begin{itemize}
    \item In Section \ref{sec:intro}, we defined the problem of multi-source causal Bayesian optimization and highlight its significance in real-world applications.  
    \item In Section \ref{sec:methods}, we develop multi-source causal Bayesian optimization (MSCBO), a novel algorithm that integrates causal modeling and multi-source optimization in a principled way.  
    \item In Section \ref{sec:results}, we validate MSCBO through extensive experiments on synthetic and real-world benchmarks, showcasing its superior performance compared to existing methods.  
\end{itemize}

By addressing the dual challenges of causal and multi-source optimization, MSCBO represents a significant step forward in the field of Bayesian optimization, offering a cost-efficient and scalable solution for complex decision-making problems.

\section{Methodology}\label{sec:methods}

\subsection{Algorithm specification}\label{sec:algospecs}
The medicine trial example from the introduction highlights a fundamental limitation of existing approaches when applied to real-world optimization problems that involve high-dimensional decision spaces, costly interventions, and heterogeneous sources of information. Specifically, naive integrations of CBO and MSBO can result in inefficient resource allocation. On the one hand, CBO effectively leverages causal structures to identify the most impactful interventions, but it does not account for the varying fidelity or cost of information sources. On the other, MSBO efficiently integrates information from multiple sources with differing costs and accuracies but assumes an independent and identically distributed structure for each of the sources, ignoring the underlying causal relationships between variables.  This disconnect leads to two key inefficiencies: 

\begin{enumerate}
    \item Unnecessary or redundant interventions that waste budget by failing to prioritize based on causal impact.
    
    \item Over-reliance on either high-fidelity or low-fidelity sources, which may result in suboptimal solutions due to poor generalization.
    
\end{enumerate}
The combination of these challenges necessitates the design of a novel framework that integrates both causal knowledge and multi-source optimization in a principled manner, enabling efficient resource allocation and robust decision-making.

\subsection{MSCBO}

\begin{figure}[!htb]
\centering
\includegraphics[scale=0.6]{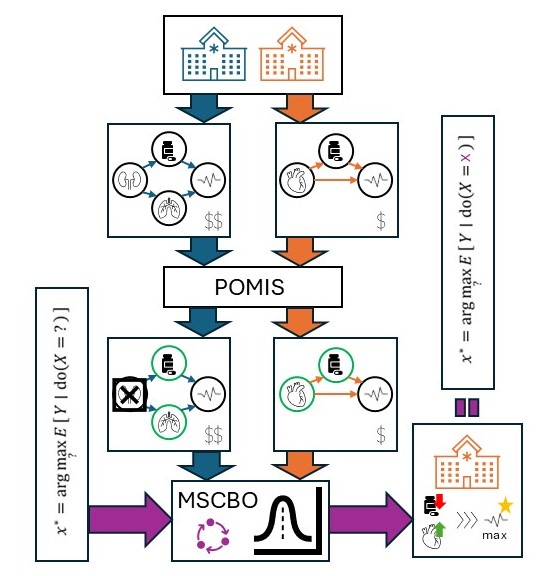}
\caption{MSCBO Conceptual Diagram: As its input the algorithm takes information from multiple sources, in the form of DAG's and their associated observation values, and an optimization query (represented by the question mark). The POMIS-algorithm serves as a subroutine to prune exploration sets. Based on these DAG's, the algorithm proposes an intervention set value assignment that maximizes or minimizes the posterior value.}
\label{fig:MSCBO}
\end{figure}

Inspired by the strengths and lessons learned from CausalBO and Multi-Source BO \citep{aglietti2020causal,roberts2024causalbo,poloczek2017multi}, we propose \textbf{MSCBO} (\textit{Multi-Source Causal Bayesian Optimization}), a novel algorithm that unifies causal knowledge and multi-information source optimization into a single framework. A conceptualized overview of MCSBO can be found in Figure \ref{fig:MSCBO}. MSCBO is designed to efficiently navigate the trade-offs between intervention cost, source fidelity, and causal impact by integrating the following key principles.  

\begin{itemize}
    \item \textbf{Causal modeling:} MSCBO explicitly incorporates causal structures to identify the most impactful interventions in the style of CBO, leveraging the sparsity of causal graphs to reduce the dimensionality of the decision space.  
    \item \textbf{Multi-source integration:} The algorithm combines information from heterogeneous sources in the style of MSBO. It balances trade-offs between cost, fidelity, and uncertainty to make decisions about which sources to query.  
    \item \textbf{Balancing exploration and exploitation:} The algorithm employs an $\epsilon$-greedy policy, a decision-making strategy that balances exploration and exploitation. 
\end{itemize}

   Specifically, MSCBO ensures that interventions are both causally impactful and cost-efficient, avoiding redundancy while achieving robust performance across heterogeneous environments. The pseudocode for the implementation can be found in Algorithm \ref{alg:pseudocode}.

\begin{algorithm}[!htb]
\caption{Multi-source Causal Bayesian Optimization (MSCBO)}
\KwIn{Initial sources \( IS_1, IS_2, \dots, IS_M \), exploration sets \( \mathcal{V}_1, \mathcal{V}_2, \dots, \mathcal{V}_M \), source fidelities \( c_1, c_2, \dots, c_M \), intervention, an objective function \(g\), observation costs \((IC,OC)\) and budget \( t\).}
\KwResult{Best source \( IS^* \), total cost \(c^*\), best intervention set \(X^*\) and corresponding value estimate \( v = g(\text{do}(X^* = x)\).}
\begin{algorithmic}[1]
\FOR{step: s}
    \FOR{source \(IS_m\)}
        \STATE Optimize and evaluate Cost-sensitive Information Gradient over the intervention set \text{CKG}(\(IS_m, c_m\))
    \ENDFOR
    \STATE Select source \(IS^*\) and intervention set \(X^* \subseteq \mathcal{V}\) with maximal Information Gradient value \( \{IS^*,X^*\} = \max_{m \in \{1, \dots, M\}} \text{CKG}(IS_m,\mathcal{V}_m, c_m) \).

    \STATE Calculate \( \epsilon \)-greedy policy value \( \epsilon(IS^*)\).

    \IF{\( \epsilon (IS^*) > \theta \sim \mathcal{U}(0,1) \)}
        \STATE Collect \(k\) new observational samples from \(IS^*\).
        \STATE Update the causal \(GP\) model of the optimal source \(IS^*\) with the new observational samples.
        \STATE Add the observation cost multiplied with the length of the intervention set to the total cost \(c^* = c^* + OC\ \cdot\ \text{length}(X^*)\).
    \ELSE
        \STATE Collect the best set of variables to intervene on \(X^*\), and its optimal value assignment \(\text{do}(X^* = x) \).
        \STATE Perform the intervention and update the causal \(GP\) model of the optimal source \(IS^*\) and the current global optimum \(v = g(\text{do}(X^* = x)\) accordingly.
        \STATE Add the intervention cost multiplied with the length of the intervention set to the total cost \(c^* = c^* + IC\ \cdot\ \text{length}(X^*) \).
    \ENDIF
    \IF{\(c^* \ge t\)}
        \STATE Terminate, return the result variables.
    \ENDIF
\ENDFOR
\end{algorithmic}
\label{alg:pseudocode}
\end{algorithm}

The algorithm requires the user to provide the information sources (the DAG's and associated observations), their costs, an intervention- and observation cost, a ground truth (for performance evaluation) and a general budget for interventions/observations. Based on these variables, the algorithm will commence execution. It operates in iterations, working within the constraints of the budget. Each iteration is structured as follows.

\begin{enumerate}
\item Optimize and Evaluate the Cost-sensitive information Gradient (CKG), outlined in Section \ref{sec:CKG}, for each of the information sources (2,3).
\item Collect the source producing the maximum CKG-value (5).
\item Determine whether to intervene or observe based on the \( \epsilon \)-greedy policy as presented in Section \ref{sec:egreedy} (6,7).
\begin{enumerate}
    \item \textbf{IF} observe, collect new observational data from the previously collected best source's network (\(k\) samples, where \(k\) is user defined) and update the hyperparameters of its causal model with this data (8,9).\
    \item \textbf{IF} intervene, collect the optimal intervention set from the previously collected best source, perform the intervention to compute a posterior value, update the hyperparameters of its causal model and update the global optimum if the existing optimum is exceeded by the computed posterior value (12,13).
\end{enumerate}
\item Update running cost, terminate operation if the total budget is exceeded and repeat step 1 if the budget is not exceeded [10,14].
\end{enumerate}

After completion of the algorithmic process, one is left with the optimal intervention set - value pair along with the information source that the pair originates from. The process of optimization over all sources, as well as performing the actual interventions, has been parallelized with the goal of decreasing the overall computation time. A repository containing the code and data can be found at \url{https://github.com/LuukJacobs1/MSCBO.git}.

\subsubsection{The Knowledge Gradient}\label{sec:CKG}

A slightly adapted cost-sensitive multi-source Knowledge Gradient, as proposed in \citet{poloczek2017multi}, forms our acquisition function. Formally, it is defined as:
\vspace{5mm}
\begin{center}
\small
\(
\max_{is \in IS} \mathbb{E}_{i} \Biggl[ \frac{\max_{x' \in \mathcal{D}_{is}} \mu^{(i+1)}(is, x') - \max_{x' \in \mathcal{D}_{is}} \mu^{(i)}(is, x')}{c_{is}} \Biggl|
IS^{(i+1)} = is, x^{(i+1)} = x \Biggr],
\)
\end{center}

\vspace{5mm}

\normalsize

\noindent where:
\begin{itemize}
    \item \(\max_{x' \in \mathcal{D}_{is}} \mu^{(i)}(is, x')\): The current running optimum for source \(is\).
    \item \(x\ \text{: The value assignment}\  \text{do}(X=x) \) to source \(is\).
\end{itemize}

As opposed to optimizing a single acquisition function over all information sources, we assign a separate Gaussian Process and acquisition function to each of the sources. This choice is mainly motivated by the desire to allow for heterogeneity in the exploration sets (e.g., the ability for the exploration sets of two sources to contain different variables). Unlike heuristics like Expected Improvement (EI) or Upper Confidence Bound (UCB), which are based on proxies for good decisions, the CKG directly evaluates the expected gain in future decision quality. It selects points that maximize the expected improvement in the best estimate of the objective, rather than just maximizing immediate gains, which is why it becomes especially useful in noisy optimization.

\subsubsection{Minimal Intervention Sets}\label{sec:pomis}

In many cases, desirable exploration sets for CBO appear as a subset of the set of intervenable nodes. These sets are referred to as Posterior Optimal Minimal Intervention Sets (POMIS). A POMIS typically encompasses nodes that have a stronger relationship to the query variable from a causal standpoint. It is formally defined as the intervention set \( I \) and corresponding values \( x_I^* \) that satisfy:
\begin{itemize}
    \item Minimality: \( I \) is the smallest set of variables such that no proper subset \( J \subset I \) can achieve the same maximum for the objective function \( g(x) \).
    \item Optimality: The intervention values \( x_I^* \) maximize the expected value of \( g(x) \) under the causal model:
    \[
    x_I^* = \arg \max_{x_I} \mathbb{E}[g(X_{\text{do}(x_I)})].
    \]
\end{itemize}

The algorithm used to construct the POMIS's has been sampled directly from the \citep{lee2018structural} paper on multi-armed bandits and thus we refer to their paper for the background theory on the determination of candidate sets. As a tiebreaker for sets of equal size, we include a summed distance of the nodes within the POMIS to the output node. An advantage of the POMIS-usage is improved scalability. Compared to MSBO, which operates on the set of all intervenable nodes, the causal approach offers better performance on larger networks, as it only targets minimal sets that achieve a similar posterior effect. Thus, the optimization problem involves fewer dimensions, significantly reducing computational overhead. This becomes a larger factor in more complex problems, where the number of observable extraneous variables is typically also significantly larger.

\subsubsection{The Cost of Intervening}
To quantify the ramifications of interventions and observations, MSCBO incorporates both an intervention and an observation cost. In its base case, this intervention cost is uniform across the variable space, and an adequate value that is proportional to the budget may be defined by an expert within the problem domain. The observation follows a similar concept, as it may also be defined by a domain expert and is uniform across the variable space. Observations are typically less costly and resource-intensive than interventions, which should be taken into account when adjusting the pair.

\subsubsection{Exploration versus Exploitation}\label{sec:egreedy}
To balance the exploration-exploitation trade-off, MSCBO incorporates an \( \epsilon \)-greedy policy that guides the algorithmic decisions. The policy is formally defined as follows:
\vspace{2mm}

\noindent Let:
\begin{itemize}
    \item \( \mathcal{X} \subseteq \mathbb{R}^d \): The space of interventional variables, defined by the bounds in \( \text{interventional domain} \),
    \item \( D_{\text{obs}} = \{x_1, x_2, \dots, x_n \} \subset \mathcal{X} \): The observational samples that satisfy the bounds of \( \mathcal{X} \),
    \item \( \text{Conv}(\cdot) \): The volume of the convex hull of a set of points.
\end{itemize}

\noindent Then, \( \epsilon \) is defined as:
\[
\epsilon = 
\begin{cases} 
\frac{\text{Conv}(D_{\text{obs}})}{\text{Conv}(\mathcal{X})} \cdot \frac{n}{n_{\text{max}}}, & \text{if } D_{\text{obs}} \neq \emptyset, \\
1, & \text{if } D_{\text{obs}} = \emptyset,
\end{cases}
\]
where:
\begin{itemize}
    \item \( \text{Conv}(D_{\text{obs}}) \): The volume of the convex hull of the observational samples \( D_{\text{obs}} \),
    \item \( \text{Conv}(\mathcal{X}) \): The volume of the convex hull defined by the bounds of the interventional domain,
    \item \(n\): The number of samples in \(D_{\text{obs}}\), and \( n_{\text{max}} \): The maximum number of samples.
\end{itemize}

\noindent After calculation of the \( \epsilon \)-value, we compare this value to a number sampled from a uniform distribution between 0 and 1, and intervene whenever the \( \epsilon \)-value exceeds this randomly sampled value. In all other cases, we abstain from intervening and instead gather new observations.
 
\subsection{Experimental Setup}
\begin{figure}
\centering
\includegraphics[scale = 0.45]{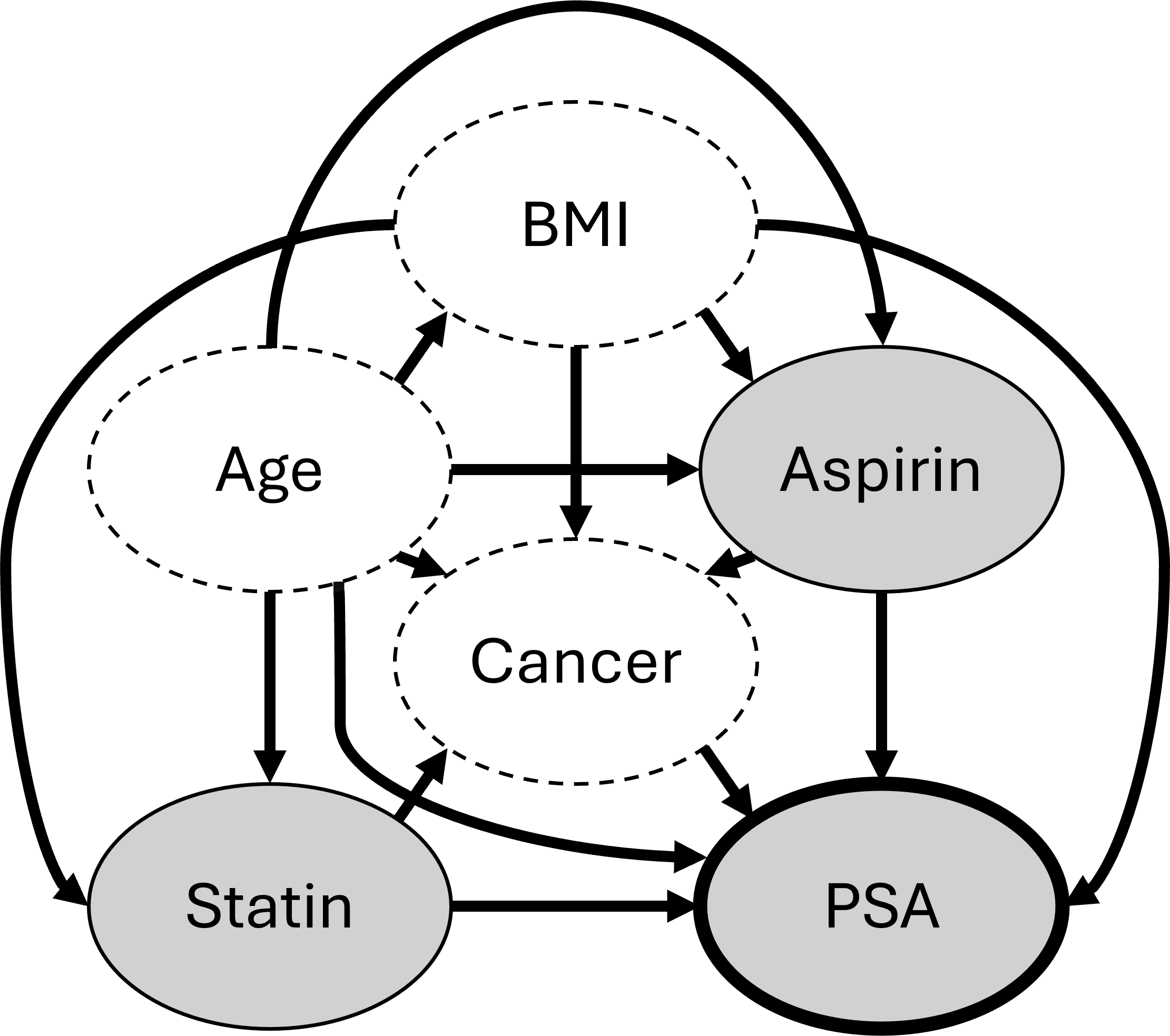}
\caption{Causal DAG depicting the PSA scenario. Gray nodes represent variables that can be intervened upon, and dashed nodes represent non-manipulative variables, respectively. The output variable PSA is denoted with a thick-dashed node.}
\label{fig:Psa}
\end{figure}

To accommodate for rigorous testing of the MSCBO-algorithm, the implementation has been applied to a mix of Bayesian Network repository networks and networks directly sampled from papers within the field of Bayesian Optimization. For clarity purposes, the results section will only discuss a causal network constructed from research on the effect of statins’ (a medicine group) use on Prostate Specific Antigen (PSA) levels \citep{ferro2015use,thompson2019causal} and a causal network representing the regulatory interactions among 46 genes in E. coli, as constructed by \citet{schafer2005shrinkage}, sampled from the Bayesian Network Repository \citep{bnrepository}. The results and structural equations belonging to the other test networks are reported in Appendix \ref{appendix:simresults}. A conceptual overview of the PSA network can be found in Figure \ref{fig:Psa}, and its structural equations are defined as follows. 

\[
\begin{aligned}
\text{age} &= \mathcal{U}(55, 75), \\
\text{bmi} &= \mathcal{N}(27.0 - 0.01 \cdot \text{age},0.7) \\
\text{aspirin} &= \sigma(-8.0 + 0.1 \cdot \text{age} + 0.03 \cdot \text{bmi}) \\
\text{statin} &= \sigma(-13.0 + 0.1 \cdot \text{age} + 0.2 \cdot \text{bmi}) \\
\text{cancer} &= \sigma(2.2 - 0.05 \cdot \text{age} + 0.01 \cdot \text{bmi} - 0.04 \cdot \text{statin} + 0.02 \cdot \text{aspirin}) \\
\text{psa} &=
\mathcal{N}\Big(\!6.8 + 0.04 \cdot \text{age} - 0.15 \cdot \text{bmi} - 0.6 \cdot \text{statin} \\
&\quad + 0.55 \cdot \text{aspirin} + 1.00 \cdot \text{cancer}, 0.4)
\end{aligned}
\]
\vspace{1mm}

Due to its large size, we refer to Appendix \ref{appendix:ecoligraphoverview} for the structural overview of the E. coli network and to the Bayesian Network Repository for its structural equations. These instances were chosen due to their real-world validity and both the complexity and non-linearity of the causal relationships between the network variables. 
In the PSA example, we use the predefined output node \(\textbf{PSA}\) and exploration set \( \{ \textbf{Aspirin}, \textbf{Statin} \} \) and define an optimization problem with the goal of minimizing \(\textbf{PSA}\). According to Aglietti's \citep{aglietti2020causal} findings, the best possible intervention is \((X^*, x^*) = (\{ \textbf{Aspirin}, \textbf{Statin} \}, (0.0, 1.0))\).

For the E. coli example, we define the gene \(\textbf{b1583}\) as our output node and define an optimization problem with the goal of minimizing \(\textbf{b1583}\). To keep the computation tractable, we consider intervention in all ancestors of \(\textbf{b1583}\), as opposed to intervention on the entire graph. This ancestral set consists of the nodes \(\{\textbf{lacZ},\textbf{lacY},\textbf{lacA},\textbf{asnA},\textbf{cspG},\textbf{eutG},\textbf{ygcE},\textbf{sucA},\textbf{yceP}\}\) and thus forms our exploration set \(\mathcal{V}\). All other network nodes are considered to be observable but non-intervenable. Since the authors of the paper do not provide an empirically found optimal intervention and manual computation is intractable, we cannot assess whether the algorithms actually converge to the global optimum, much like a real-world scenario. However, in such cases we can still inspect the comparative performance over cost of MSCBO, CBO and MSBO. 

\subsection{Experiment Parameters}{\label{sec:params}}
The experimental setup for these networks encompasses a range of three testing scenarios. This setup serves the purpose of assessing the robustness and scalability of our implementation while also determining its comparative performance to both a non-causal multi-source and a causal single-source approach. Each of the three scenarios mirrors a real-world case of missing/noisy information, with the gradation increasing from scenario one to three. As a base case, we will consider the networks as defined in their corresponding papers/repositories. 

\begin{enumerate}
    \item Changes in the structural equation models that DO NOT alter the connections in the network.
    \item Changes in the structural equation models that DO alter the connections in the network.
    \item Changes in the structural equation models that remove existing nodes from the network.
\end{enumerate}

In each of the scenarios, a unit cost has been defined that represents the amount of intervention/observation work done by the algorithm. This cost metric is directly inherited from the CausalBO package \citep{roberts2024causalbo}. For simplicity, each intervention is assigned an equal cost of 20 units and each observation is also assigned an equal cost of 1 unit. Each scenario makes use of two independently generated sources. Querying costs are set to be uniform among the information sources and their variables in order to assess the raw comparative performance in a basal scenario. Performance is measured by running each algorithm with ten separate GP prior initializations and collecting the optima and cost for each algorithmic step. We report the performance over cost in plots and the optima found after convergence in tables. An implementation that either provides better cost-effectiveness or manages to reach a better optimum value is considered superior to its competitors.

\begin{figure*}[htb]
    \centering
    \begin{subfigure}{0.45\textwidth}
        \includegraphics[width=\linewidth]{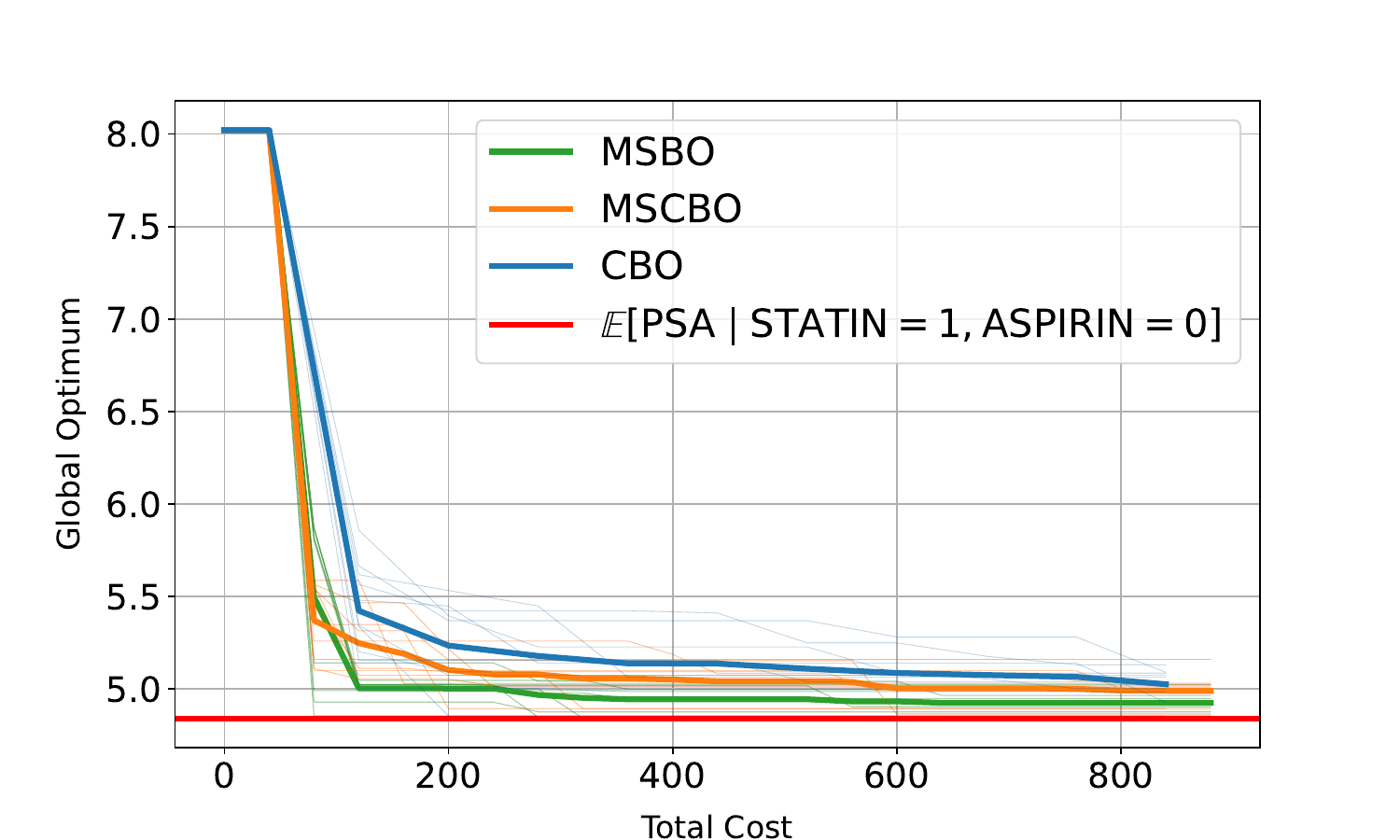}
        \caption{\textbf{The Base Case}}
    \end{subfigure}
    \hspace{0.05cm}
    \begin{subfigure}{0.45\textwidth}
        \includegraphics[width=\linewidth]{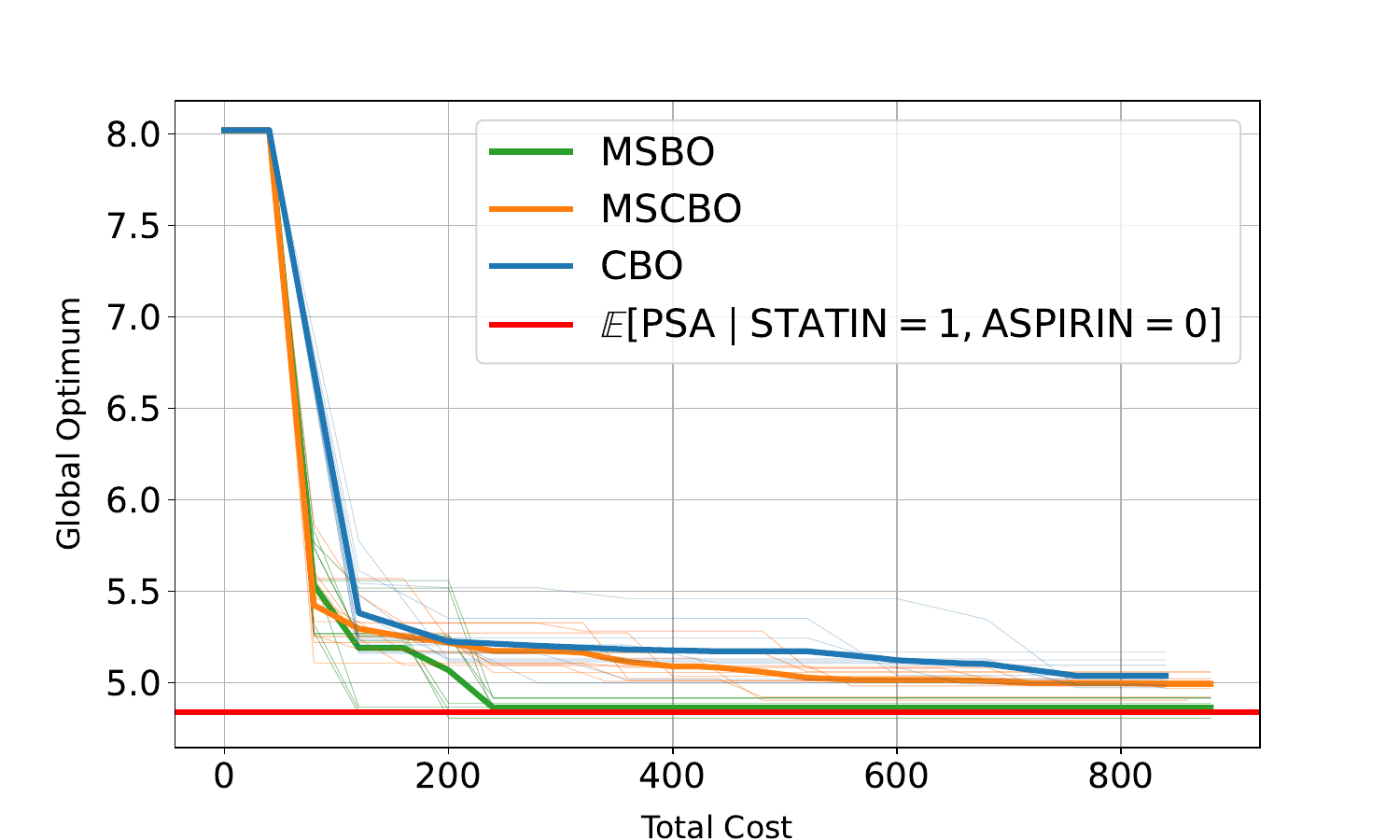}
        \caption{\textbf{Scenario 1}: Altered SCM's}
    \end{subfigure}

    \vspace{-0.12cm}

    \begin{subfigure}{0.45\textwidth}
        \includegraphics[width=\linewidth]{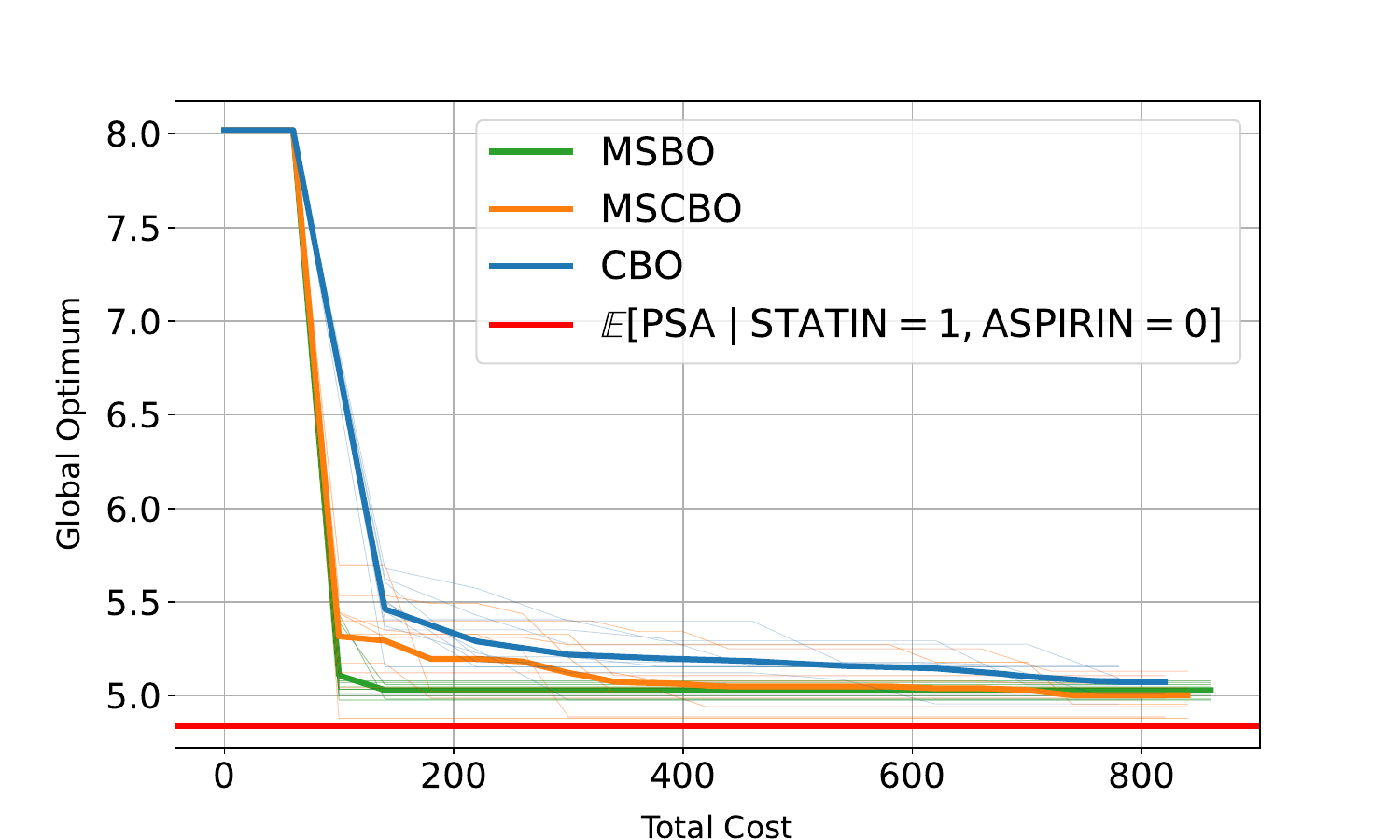}
        \caption{\textbf{Scenario 2}: Added/removed node connections}
    \end{subfigure}
    \hspace{0.05cm}
    \begin{subfigure}{0.45\textwidth}
        \includegraphics[width=\linewidth]{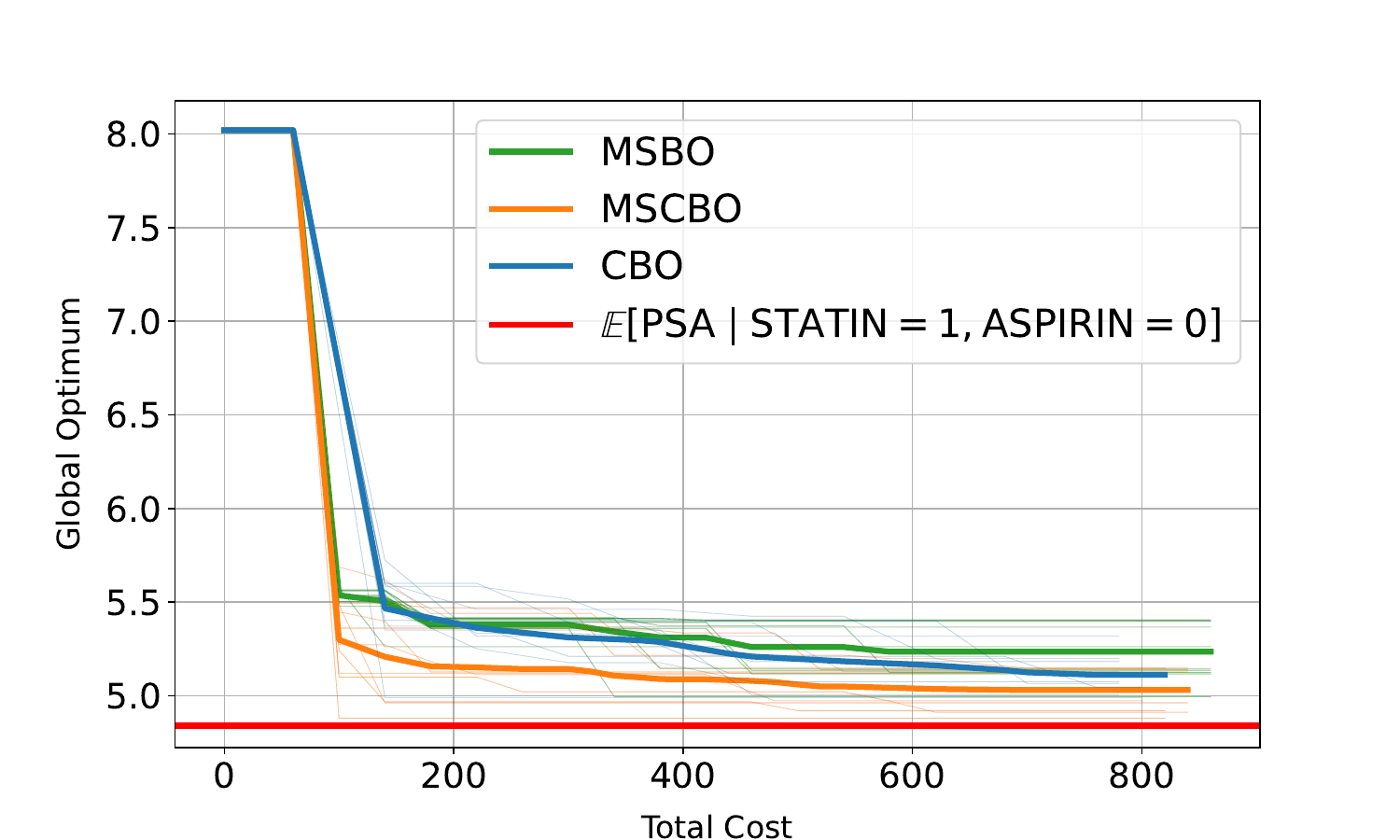}
        \caption{\textbf{Scenario 3}: Added/removed nodes}
    \end{subfigure}

    \caption{Optima over cost for the three different implementations within the PSA graph. The true optimum is represented by the red line in the graph.}
    \label{fig:psaresults}
\end{figure*}

\section{Results}\label{sec:results}

\subsection{PSA Example}\label{sec:PSA_res}
For the PSA example, the optimal output value is denoted as:   \[
\mathbb{E}\!\left[\, \text{PSA} \;\middle|\; \text{do}(\textbf{Aspirin}=0.0, \;\textbf{Statin}=1.0) \,\right]
\approx 4.8\] 
As can be observed in Figure \ref{fig:psaresults}, average cost-efficiency of MSCBO is comparable to, or slightly exceeds, that of single-source CBO in all scenarios. This is the direct result of its smart source selection, which allows for single-source intervention while exploring the solution space of both sources, thus halving intervention costs. The global optimum is roughly approximated in three of the four test cases. Performance is slightly worse cost-wise compared to MSBO and this can partially be attributed to the \( \epsilon \)-greedy policy always observing in the first algorithmic iteration. In addition, the slight discrepancy might be the result of the causal kernel and DoWhy causal effect estimation, which has a tendency to be unreliable in certain areas of the interventional space \citep{roberts2024causalbo}. Since this network does not allow for POMIS-based exploration set pruning, MSCBO does not have a concrete comparative advantage over MSBO. Both intervene on the same set of variables \( \{ \textbf{Aspirin}, \textbf{Statin} \} \) in a single source for each iteration. In scenario 3, where noise has been introduced in the form of removed nodes, we observe a decline in performance for MSBO. In contrast, both CBO and MSCBO demonstrate better robustness, indicating that the use of causal information offsets the introduced noise.

These results indicate that the optimum-finding performance of MSCBO is reasonably interchangeable with that of MSBO while the cost efficiency appears to be slightly worse. However, in the presence of significant noise, the performance advantage appears to shift to the causal implementation. In addition, one must consider that MSCBO is employed in its "weakest" context, given that no exploration set pruning is possible. In essence, this example serves as a representation of baseline performance, demonstrating that MSCBO can compete with both of its foundational counterparts.

\begin{figure*}[htb]
    \centering
    \begin{subfigure}{0.45\textwidth}
        \includegraphics[width=\linewidth]{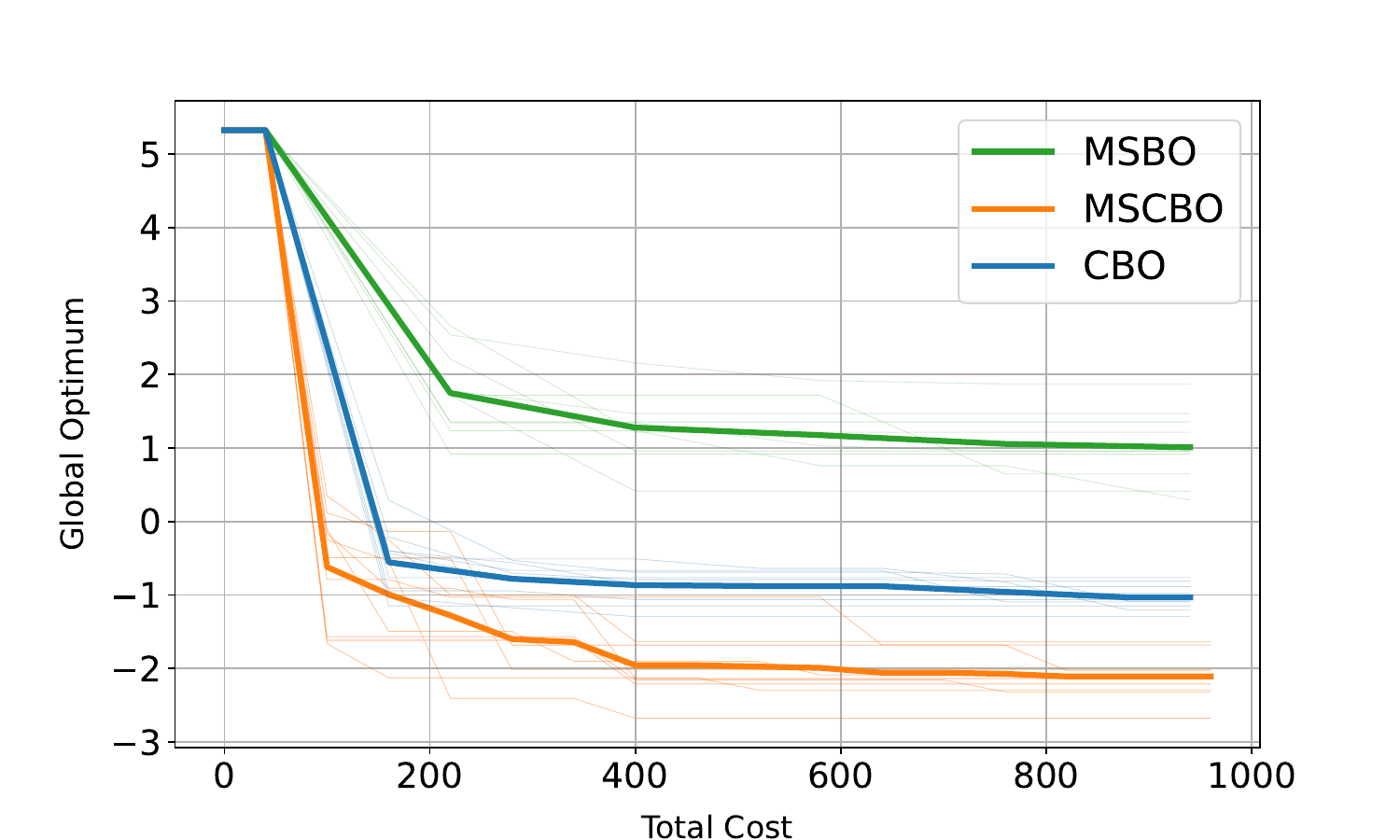}
        \caption{\textbf{The Base Case}}
    \end{subfigure}
    \hspace{0.05cm}
    \begin{subfigure}{0.45\textwidth}
        \includegraphics[width=\linewidth]{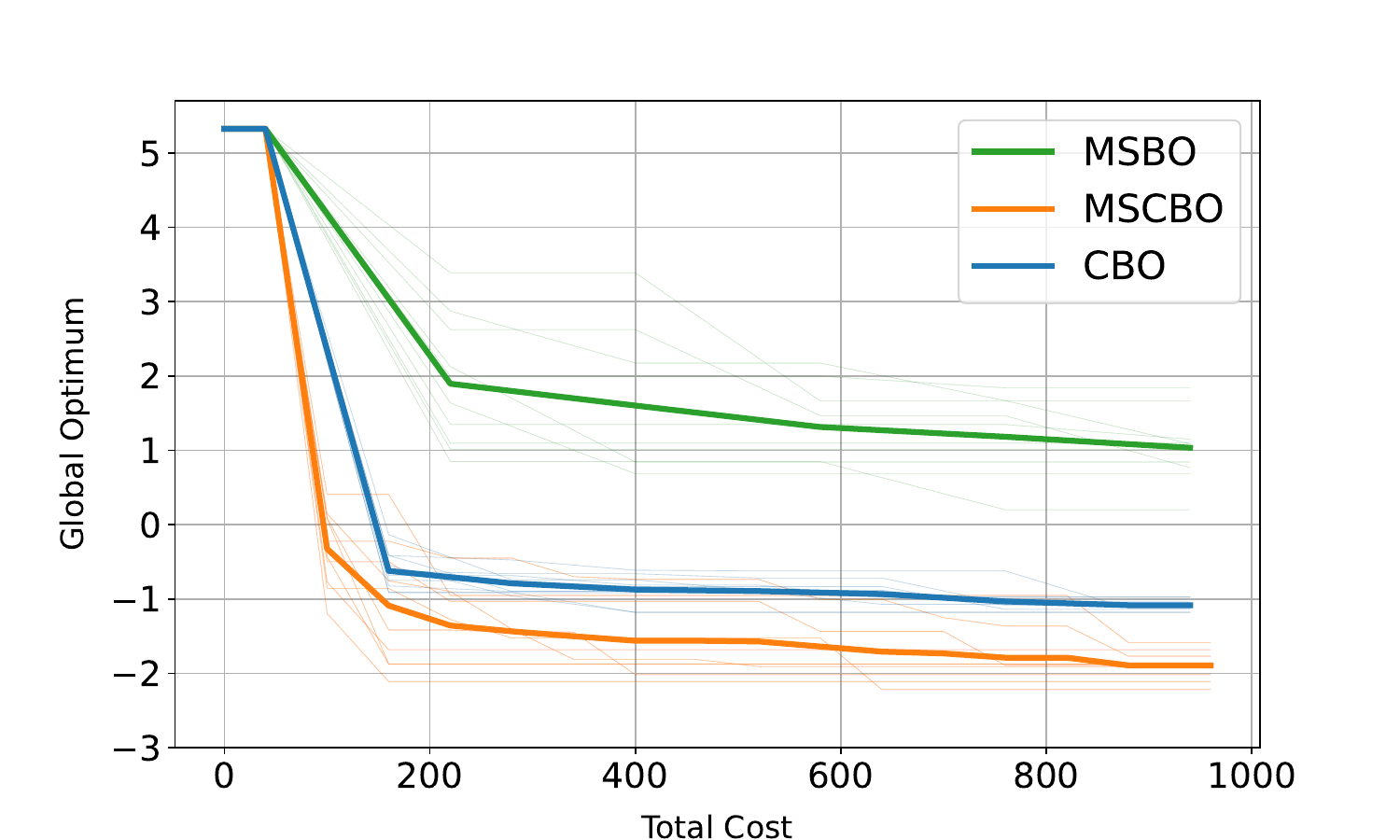}
        \caption{\textbf{Scenario 1}: altered SCM's}
    \end{subfigure}

    \vspace{-0.12cm}
    
    \begin{subfigure}{0.45\textwidth}
        \includegraphics[width=\linewidth]{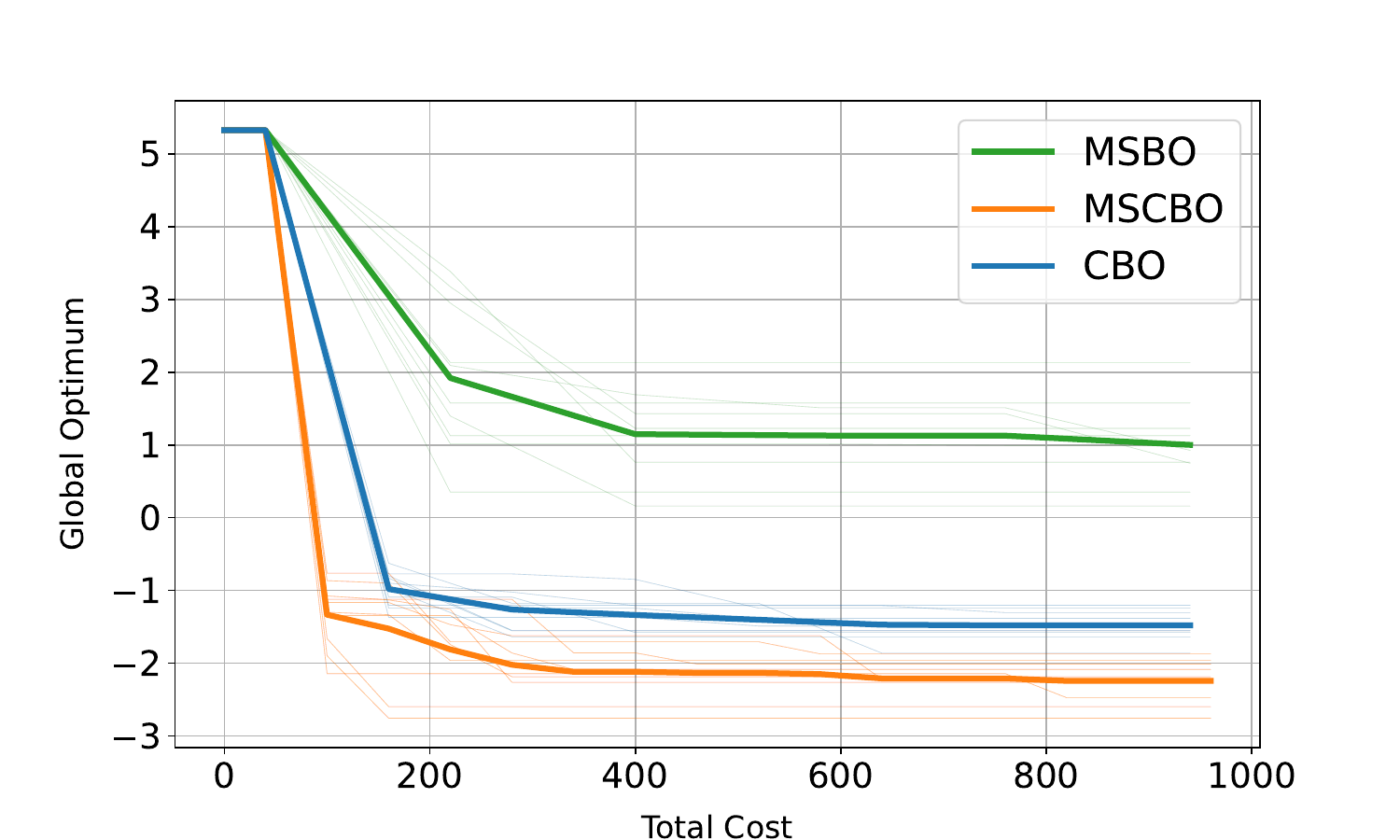}
        \caption{\textbf{Scenario 2}: Added/removed node connections}
    \end{subfigure}
    \hspace{0.05cm}
    \begin{subfigure}{0.45\textwidth}
        \includegraphics[width=\linewidth]{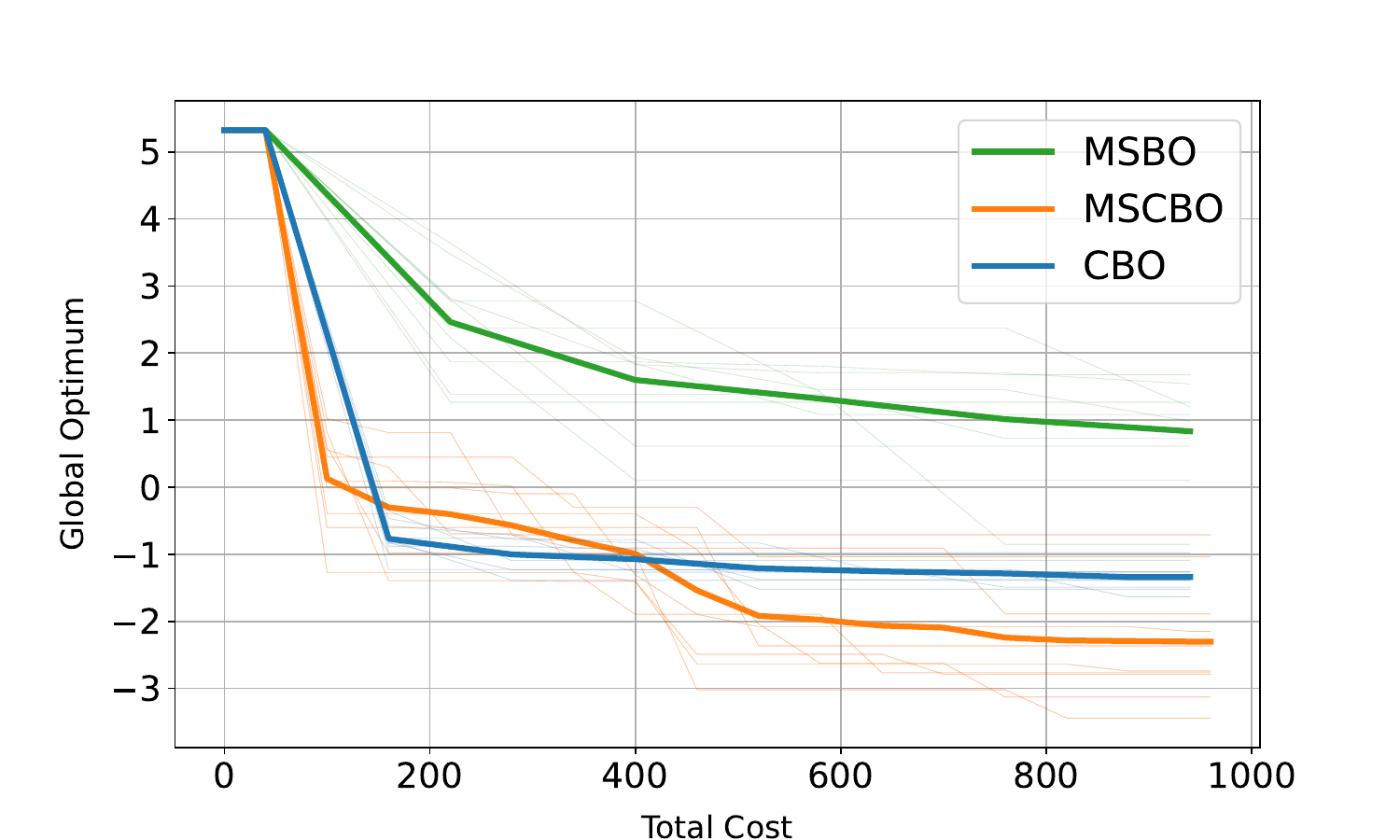}
        \caption{\textbf{Scenario 3}: Added/removed nodes}
    \end{subfigure}

    \caption{Optima over cost for the three different implementations within the E. coli graph.}
    \label{fig:ecoliresults}
\end{figure*}



\subsection{E-coli Example}
Due to the intractability of posterior optimization in high dimensional scenarios, an optimal intervention cannot feasibly be calculated for the E. Coli network. The results for each of the four scenarios can be observed in Figure \ref{fig:ecoliresults}. The results highlight the improved scalability of causality-based implementations over standard BO in large network variants. Both MSCBO and CBO strongly outperform MSBO in cost efficiency in all scenarios. This can be attributed to the exploration set pruning achieved by implementing the POMIS-subroutine. Doing so creates a search space that is significantly smaller than that of MSBO. More specifically, we cut down from an exploration set of \(\{\textbf{lacZ},\allowbreak \textbf{lacY},\allowbreak \textbf{lacA},\allowbreak 
\textbf{asnA},\allowbreak \textbf{cspG},\allowbreak \textbf{eutG},\allowbreak 
\textbf{ygcE},\allowbreak \textbf{sucA},\allowbreak \textbf{yceP}\}\) to an exploration set of \(\{\textbf{lacZ},\textbf{lacA},\textbf{yceP}\}\), and thus new optima are found at a significantly lower cost for both MSCBO and CBO. In practice, this results in a significant performance boost which is especially noticeable in large networks such as the E.coli network. As can be observed from Figure \ref{fig:ecoliresults}, MSCBO generally reports better cost-efficiency and convergence values compared to CBO. This can be attributed to the single source intervention rendering cheaper interventions that produce equal or better optima. In terms of robustness, both MSCBO and CBO achieve generally stable performance. Slight drop-offs in the average of found optima can be observed in scenarios 1 and 3, and are likely the result of the algorithms struggling to model a GP that accurately models the ground truth under increased noise pressure. MSBO does not appear to demonstrate any specific performance drop-off, and we hypothesize that this is the result of the algorithm getting stuck in local and easy-to-reach optima. Adding noise does not seem to influence this process. 

Overall, in the worst case, as observed within the PSA graph scenarios, the performance of MSCBO is comparable to the next-best alternative, MSBO. If we play into the strengths of MSCBO, as observed within the E. coli graph scenarios, performance exceeds both alternatives. Furthermore, the comparative advantage of MSCBO is expected to increase with the addition of more and larger sources. Where MSBO and CBO will observe a significant growth in interventions costs for larger networks and more sources respectively, MSCBO is better equipped to navigate the higher dimensionality due to its combination of exploration set pruning and smart source selection. Results for the graphs tested within the supplementary results paint a very similar picture, and thus we conclude that MSCBO presents itself as a superior optimization algorithm in multi-source Bayesian optimization scenarios where causal information is available.

\vspace{-0.2em}
\section{Discussion}
Experimental results highlight the potential of MSCBO to extend existing optimization frameworks, especially those that operate on causal processes. In a worst-case scenario, the performance of MSCBO is comparable to that of its standard BO counterpart, especially when significant noise is introduced. However, in problem settings that allow for exploration set reduction and leveraging of causal information, its features produce faster convergence, better budget handling and more accurate predictions. These results are consistent with those reported in single-source settings \citep{aglietti2020causal} and highlight how harnessing the causality principles may also cut costs in multi-source settings. MSCBO provides an efficient integration of CBO and MSBO, consequently presenting itself as a cost-efficient alternative for both in this multi-source domain. 

We believe MSCBO has a role to play in furthering our understanding of BO problems that operate on causal dependencies. Our research has proven the capabilities of combining preexisting knowledge of causal integration and multi-source optimization, rendering an implementation that outperforms its individual parts. With the use of these new strategies, costly interventions can be omitted, allowing for more efficient problem solving. The usefulness of MSCBO spans a multitude of fields, ranging from clinical studies to insurance calculations. Looking ahead, further integration of causal knowledge with BO could help overcome the significant time complexity issues that traditional BO encounters in larger problem instances \citep{frazier2018tutorial}.

\subsection{Limitations And Recommendations}
Despite high levels of parallelization, MSCBO's integration of multiple sources accompanied with the usage of custom causal GPs introduces computational overhead, especially as the number of sources or intervention variables increases. These factors may limit scalability in high-dimensional settings. This problem could be addressed through the creation of a causal GP that blends more seamlessly with the DoWhy package. Further research may also test the validity of this implementation in the case of discrete network variations that operate on the premise of probability tables. The current setup is built on the assumption that we know the structural equations that underlie the causal relationships within the network, but discrete network variations do not offer this information. Through the use of graph mutilation and posterior prediction, one may circumvent this problem and assess the algorithm's performance in a discrete setting.

Additionally, the current algorithm approach assigns a separate GP and acquisition function to each of the information sources. This configuration limits the flow of data between different sources, but was made to allow for flexibility in terms of source structure and bounds (e.g., different sources containing variables with different interventional bounds). Further research could develop a GP and acquisition function pair that manages to house all these capabilities in a single Gaussian process in the style of misoKG \cite{poloczek2017multi}. Theoretically, this would allow for increased information flow within the acquisition function optimization, likely improving convergence time and performance along the way.

\bibliographystyle{ACM-Reference-Format} 
\bibliography{sample}

@inproceedings{mockus1974bayesian,
  author=       {Mockus, Jonas},
  title=        {On {B}ayesian methods for seeking the extremum},
  booktitle=    {Proceedings of the IFIP Technical Conference},
  pages=        {400--404},
  year=         {1974}
}

@book{mockus1989bayesian,
  title={The {B}ayesian approach to local optimization},
  author={Mockus, Jonas},
  year={1989},
  publisher={Springer}
}

@inproceedings{ranjit2019efficient,
  title={Efficient deep learning hyperparameter tuning using cloud infrastructure: Intelligent distributed hyperparameter tuning with {B}ayesian optimization in the cloud},
  author={Ranjit, Mercy Prasanna and Ganapathy, Gopinath and Sridhar, Kalaivani and Arumugham, Vikram},
  booktitle={2019 IEEE 12th international conference on cloud computing (CLOUD)},
  pages={520--522},
  year={2019},
  organization={IEEE}
}

@article{frazier2018tutorial,
  title={A tutorial on {B}ayesian optimization},
  author={Frazier, Peter I},
  journal={arXiv preprint arXiv:1807.02811},
  year={2018}
}

@inproceedings{turner2021bayesian,
  title={Bayesian optimization is superior to random search for machine learning hyperparameter tuning: Analysis of the black-box optimization challenge 2020},
  author={Turner, Ryan and Eriksson, David and McCourt, Michael and Kiili, Juha and Laaksonen, Eero and Xu, Zhen and Guyon, Isabelle},
  booktitle={NeurIPS 2020 Competition and Demonstration Track},
  pages={3--26},
  year={2021},
  organization={PMLR}
}

@article{zhan2022calibrating,
  title={Calibrating building simulation models using multi-source datasets and meta-learned {B}ayesian optimization},
  author={Zhan, Sicheng and Wichern, Gordon and Laughman, Christopher and Chong, Adrian and Chakrabarty, Ankush},
  journal={Energy and Buildings},
  volume={270},
  pages={112278},
  year={2022},
  publisher={Elsevier}
}

@article{poloczek2017multi,
  title={Multi-information source optimization},
  author={Poloczek, Matthias and Wang, Jialei and Frazier, Peter},
  journal={Advances in neural information processing systems},
  volume={30},
  year={2017}
}

@article{bareinboim2015bandits,
  title={Bandits with unobserved confounders: A causal approach},
  author={Bareinboim, Elias and Forney, Andrew and Pearl, Judea},
  journal={Advances in Neural Information Processing Systems},
  volume={28},
  year={2015}
}

@article{buesing2018woulda,
  title={Woulda, coulda, shoulda: Counterfactually-guided policy search},
  author={Buesing, Lars and Weber, Theophane and Zwols, Yori and Racaniere, Sebastien and Guez, Arthur and Lespiau, Jean-Baptiste and Heess, Nicolas},
  journal={arXiv preprint arXiv:1811.06272},
  year={2018}
}

@inproceedings{aglietti2020causal,
  title={Causal {B}ayesian optimization},
  author={Aglietti, Virginia and Lu, Xiaoyu and Paleyes, Andrei and Gonz{\'a}lez, Javier},
  booktitle={International Conference on Artificial Intelligence and Statistics},
  pages={3155--3164},
  year={2020},
  organization={PMLR}
}

@inproceedings{roberts2024causalbo,
  title={CausalBO: A {P}ython {P}ackage for {C}ausal {B}ayesian {O}ptimization},
  author={Roberts, Jeremy and Javidian, Mohammad Ali},
  booktitle={SoutheastCon 2024},
  pages={1370--1375},
  year={2024},
  organization={IEEE}
}

@book{friedman2015fundamentals,
  title={Fundamentals of clinical trials},
  author={Friedman, Lawrence M and Furberg, Curt D and DeMets, David L and Reboussin, David M and Granger, Christopher B},
  year={2015},
  publisher={Springer}
}

@book{steckler2002process,
  title={Process evaluation for public health interventions and research.},
  author={Steckler, Allan Ed and Linnan, Laura Ed},
  year={2002},
  publisher={Jossey-Bass/Wiley}
}

@article{chen2021domain,
  title={Domain adaptation under structural causal models},
  author={Chen, Yuansi and B{\"u}hlmann, Peter},
  journal={Journal of Machine Learning Research},
  volume={22},
  number={261},
  pages={1--80},
  year={2021}
}

@article{ferro2015use,
  title={Use of statins and serum levels of prostate specific antigen},
  author={Ferro, Ana and Pina, Francisco and Severo, Milton and Dias, Pedro and Botelho, Francisco and Lunet, Nuno},
  journal={Acta Urol{\'o}gica Portuguesa},
  volume={32},
  number={2},
  pages={71--77},
  year={2015},
  publisher={Elsevier}
}

@inproceedings{thompson2019causal,
  title={Causal graph analysis with the causalgraph procedure},
  author={Thompson, Clay},
  booktitle={Proceedings of SAS Global Forum},
  year={2019}
}

@article{lee2018structural,
  title={Structural causal bandits: {W}here to intervene?},
  author={Lee, Sanghack and Bareinboim, Elias},
  journal={Advances in neural information processing systems},
  volume={31},
  year={2018}
}

@article{smolinski2024causal,
  title={Causal network linking honey bee ({A}pis mellifera) winter mortality to temperature variations and {V}arroa mite density},
  author={Szymon Smoli{\'n}ski and Adam Glazaczow},
  journal={Science of The Total Environment},
  volume={954},
  pages={176245},
  year={2024},
  publisher={Elsevier}
}

@misc{bnrepository,
  author = {Gal Elidan},
  title = {{Bayesian Network Repository}},
  note ={{https://www.cse.huji.ac.il/\raisebox{-0.8ex}{\~{}}galel/Repository/}},
  year = {2001}
}

@article{schafer2005shrinkage,
  title={A shrinkage approach to large-scale covariance matrix estimation and implications for functional genomics},
  author={Sch{\"a}fer, Juliane and Strimmer, Korbinian},
  journal={Statistical applications in genetics and molecular biology},
  volume={4},
  number={1},
  year={2005},
  publisher={De Gruyter}
}

@inproceedings{li2017dropout,
  title     = {High Dimensional Bayesian Optimization using Dropout},
  author    = {Li, Cheng and Gupta, Shivaram and Rana, Santu and Nguyen, Vu and Venkatesh, Svetha},
  booktitle = {Proceedings of the 26th International Joint Conference on Artificial Intelligence (IJCAI-17)},
  pages     = {2096--2102},
  year      = {2017},
  organization = {IJCAI}
}

@inproceedings{shen2023highdim,
  title     = {Computationally Efficient High-Dimensional Bayesian Optimization via Variable Selection},
  author    = {Shen, Yihang and Kingsford, Carl},
  booktitle = {Proceedings of the Second International Conference on Automated Machine Learning (AutoML 2023)},
  series    = {PMLR},
  volume    = {224},
  pages     = {15/1--27},
  year      = {2023},
  publisher = {PMLR}
}

\newpage
\onecolumn

\title{Extending Multi-source Bayesian Optimization With Causality Principles\\(Supplementary Material)}
\maketitle

\appendix
\section{Additional simulation results}\label{appendix:simresults}
This section depicts the results of the other treated test cases and serves to support the conclusions reached within the main part of the paper. 

\subsection{Crop Yield Toy Example}
\begin{figure}[!htb]
  \centering
  \includegraphics[scale=0.4]{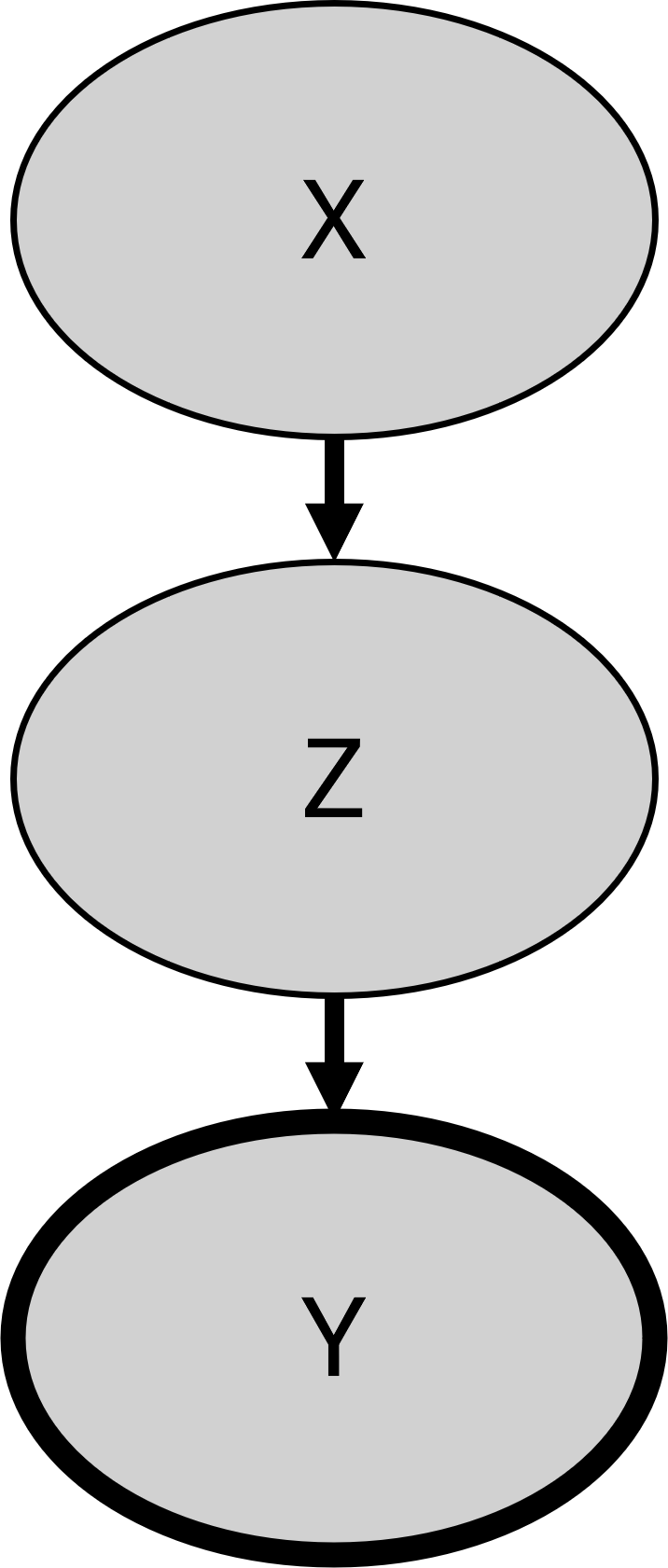}
  \caption{Causal graph depicting Crop yield. Gray nodes represent variables that can be intervened upon, and dashed nodes represent non-manipulative variables. The target variable Y is denoted with a thick-dashed node.}
 \label{fig:Crop}
\end{figure}

We treat the results of the toy example found in the  \citet{aglietti2020causal} paper. The example itself depicts the scenario of crop yield optimization.\textbf{Y} is the crop yield, \textbf{X} denotes soil fumigants and \textbf{Z} represents the eel worm population (Figure \ref{fig:Crop}). \textbf{Y} is denoted as the target variable and we define an optimization problem with the goal of minimizing \textbf{Y} and an exploration set \(\mathcal{V}\) consisting of \(\{\textbf{Z},\textbf{X}\}\) (Figure \ref{fig:Crop}). The structural equations of the graph variables are defined as follows:

\[
\begin{aligned}
X &= \epsilon_X, \\
Z &= e^{-X} + \epsilon_Z, \\
Y &= \cos(Z) - e^{-Z/20} + \epsilon_Y
\end{aligned}
\]

The MIS-optimal set consists of \(\{\textbf{Z}\}\), and the accompanying optimal intervention is defined below:

 \[
\mathbb{E}\!\left[\, \text{Y} \;\middle|\; \text{do}(\textbf{Z}=-3.2) \,\right]
\approx -2.17\] 

\begin{figure*}[htb]
    \centering
    \begin{subfigure}{0.45\textwidth}
        \includegraphics[width=\linewidth]{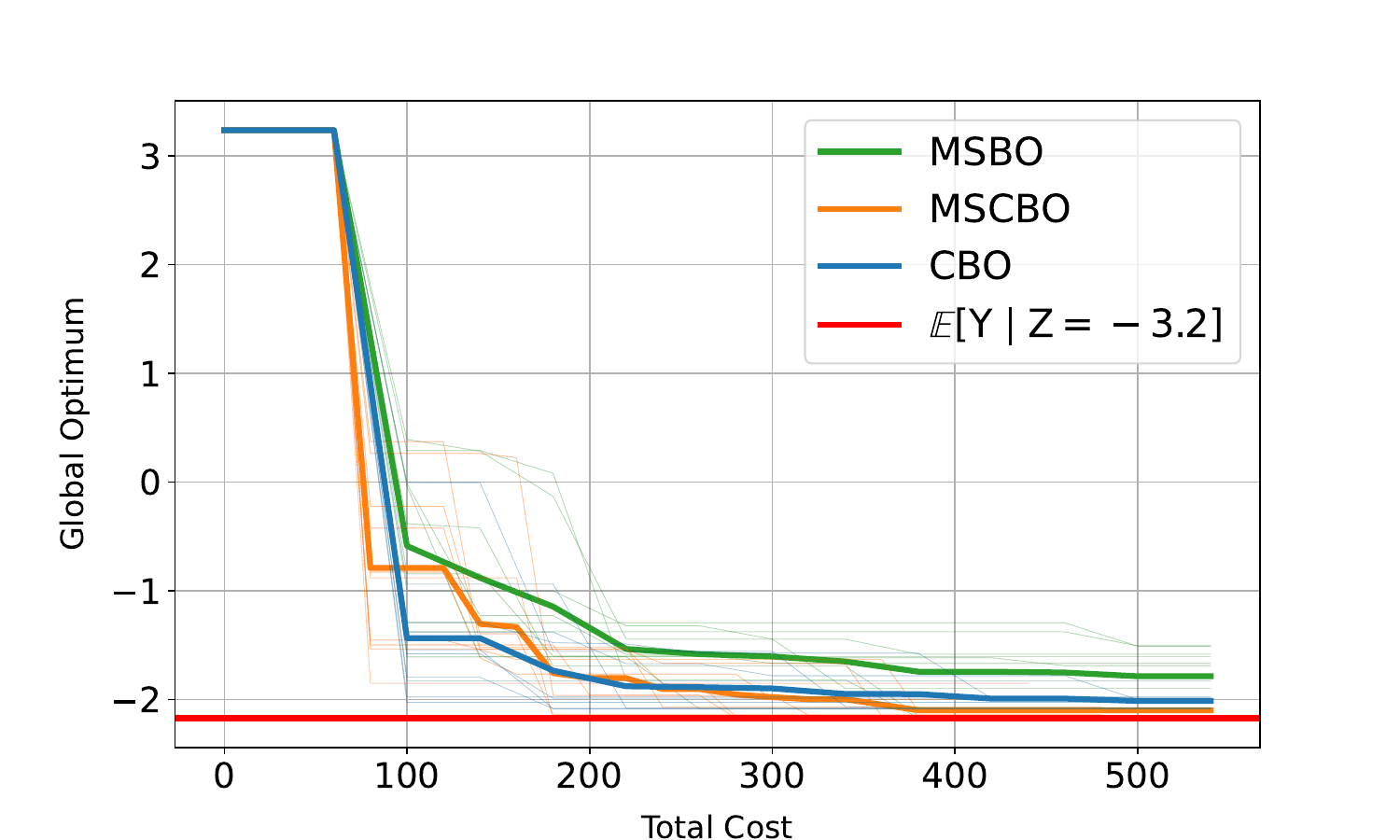}
        \vspace{0.05cm}
        \caption{\textbf{The Base Case}}
    \end{subfigure}
    \hspace{0.05cm}
    \begin{subfigure}{0.45\textwidth}
        \includegraphics[width=\linewidth]{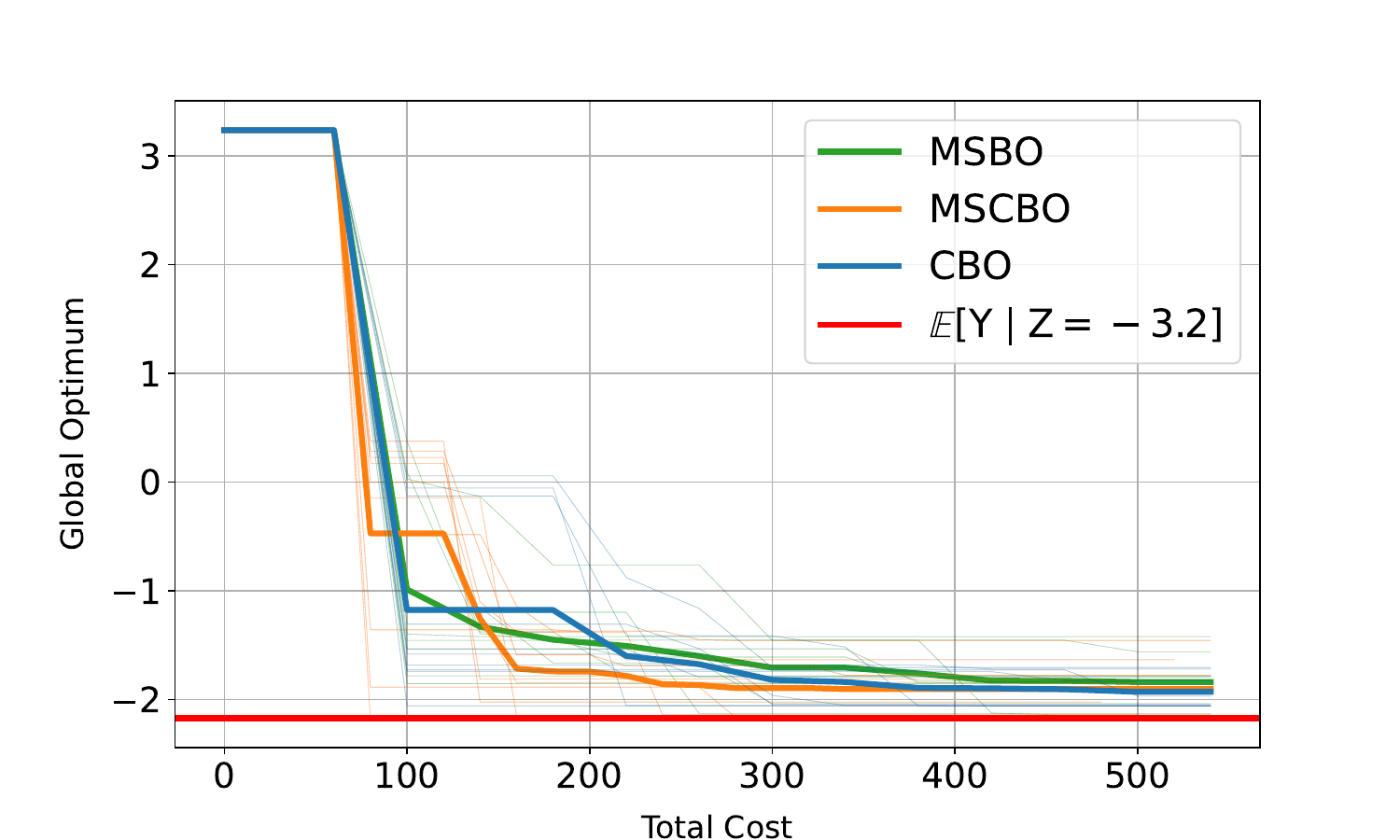}
        \vspace{0.05cm}
        \caption{\textbf{Scenario 1}: altered SCM's}
    \end{subfigure}
    
    \caption{Optima over cost for the three different implementations within the Crop Yield graph.}
    \label{fig:toyresults}
\end{figure*}

Due to the limited size and complexity of the network, we only treat the base case and scenario 1 (Figure \ref{fig:toyresults}).

The pruning of the exploration set creates an advantage for the MSCBO and CBO algorithms, as both intervene in \textbf{Z} solely, as opposed to MSBO, which intervenes in both \textbf{Z} and \textbf{X}. In the base case, performance of MSCBO and CBO exceeds that of MSBO in both found optima and cost-efficiency. All algorithms converge either on or very close to the optimal intervention on average. When noise is introduced, as observed in scenario 1, the performance of CBO and MSCBO decreases slighty and more closely resembles that of MSBO. Overall, the performance of MSCBO and CBO is very comparable, and either matches or outperforms standard MSBO in both situations. Therefore, we determine that the availability of causal knowledge allows for a better optimization process, as observed in the performance comparison of the three algorithms.

\subsection{Honey Yield Example}
\begin{figure}[!hbt]
  \centering
  \includegraphics[width=0.45\textwidth]{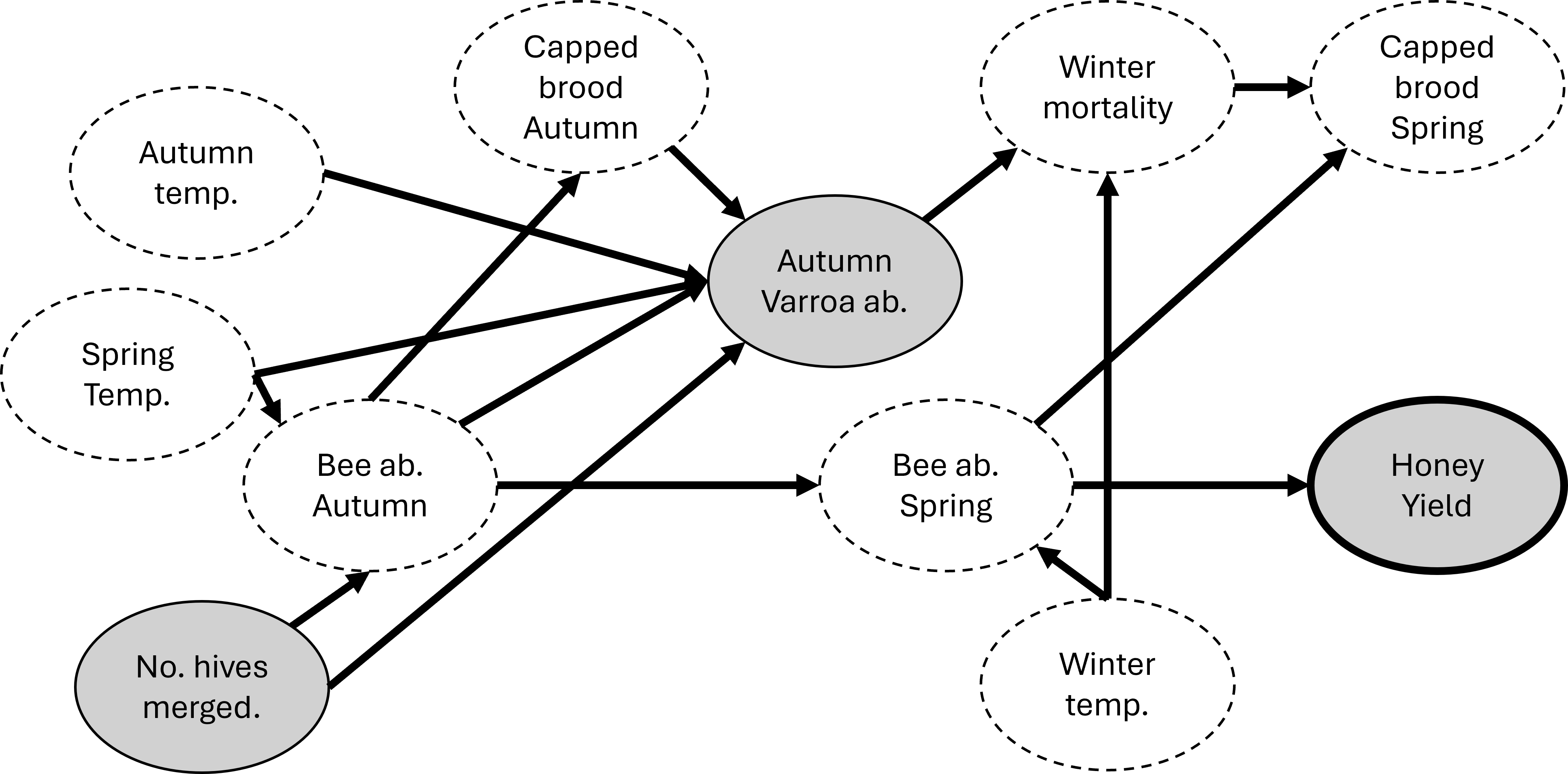}
  \caption{Causal graph depicting Honey yield. Gray nodes represent variables that can
 be intervened upon, and dashed nodes represent non-manipulative variables. The
 target variable honey yield is denoted with a thick-dashed node.}
 \label{fig:Beehive}
\end{figure}

\begin{figure*}[!htb]
    \centering
    \begin{subfigure}{0.45\textwidth}
        \includegraphics[width=\linewidth]{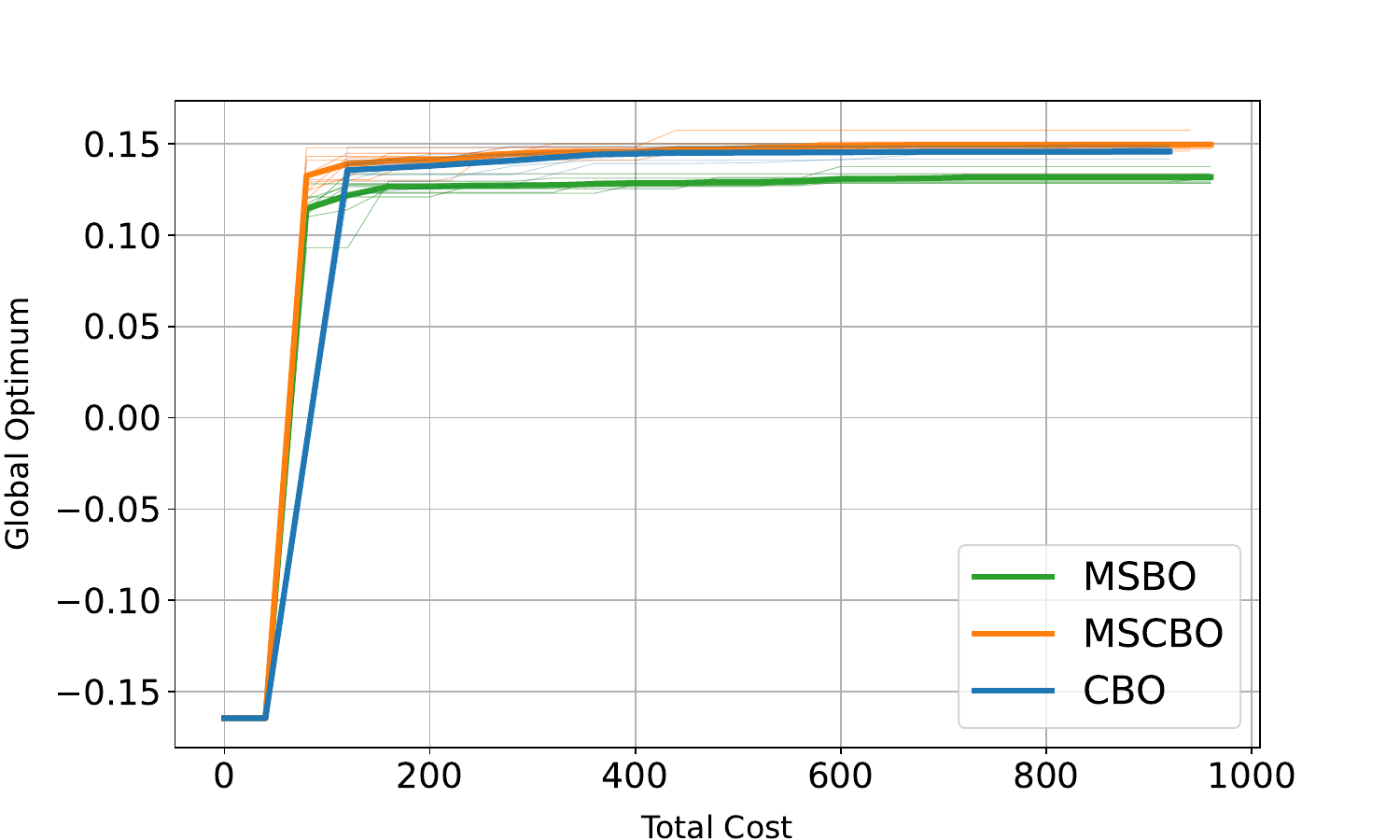}
        \caption{\textbf{The Base Case}}
    \end{subfigure}
    \hspace{0.05cm}
    \begin{subfigure}{0.45\textwidth}
        \includegraphics[width=\linewidth]{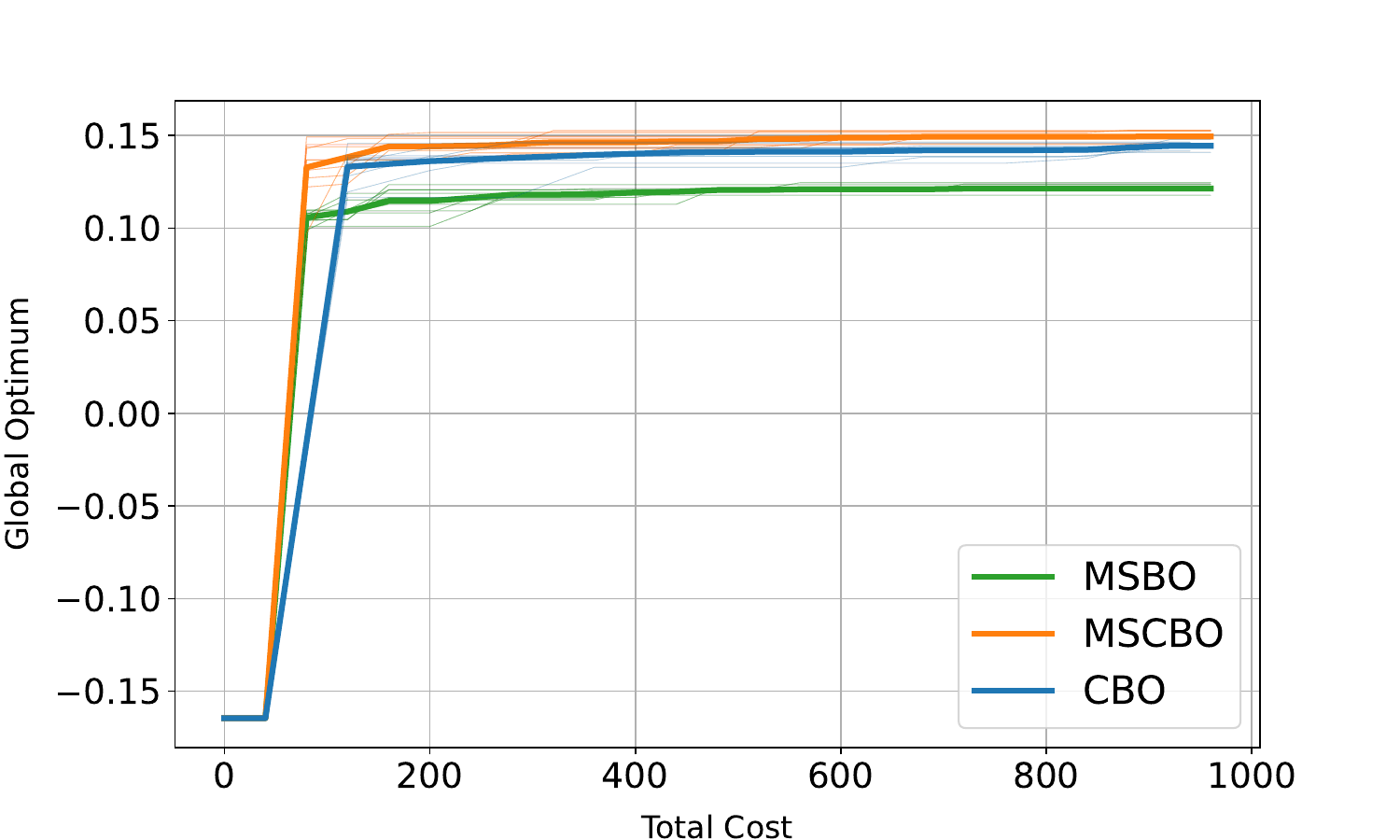}
        \caption{\textbf{Scenario 1}: altered SCM's}
    \end{subfigure}

    \vspace{-0.12cm}
    
    \begin{subfigure}{0.45\textwidth}
        \includegraphics[width=\linewidth]{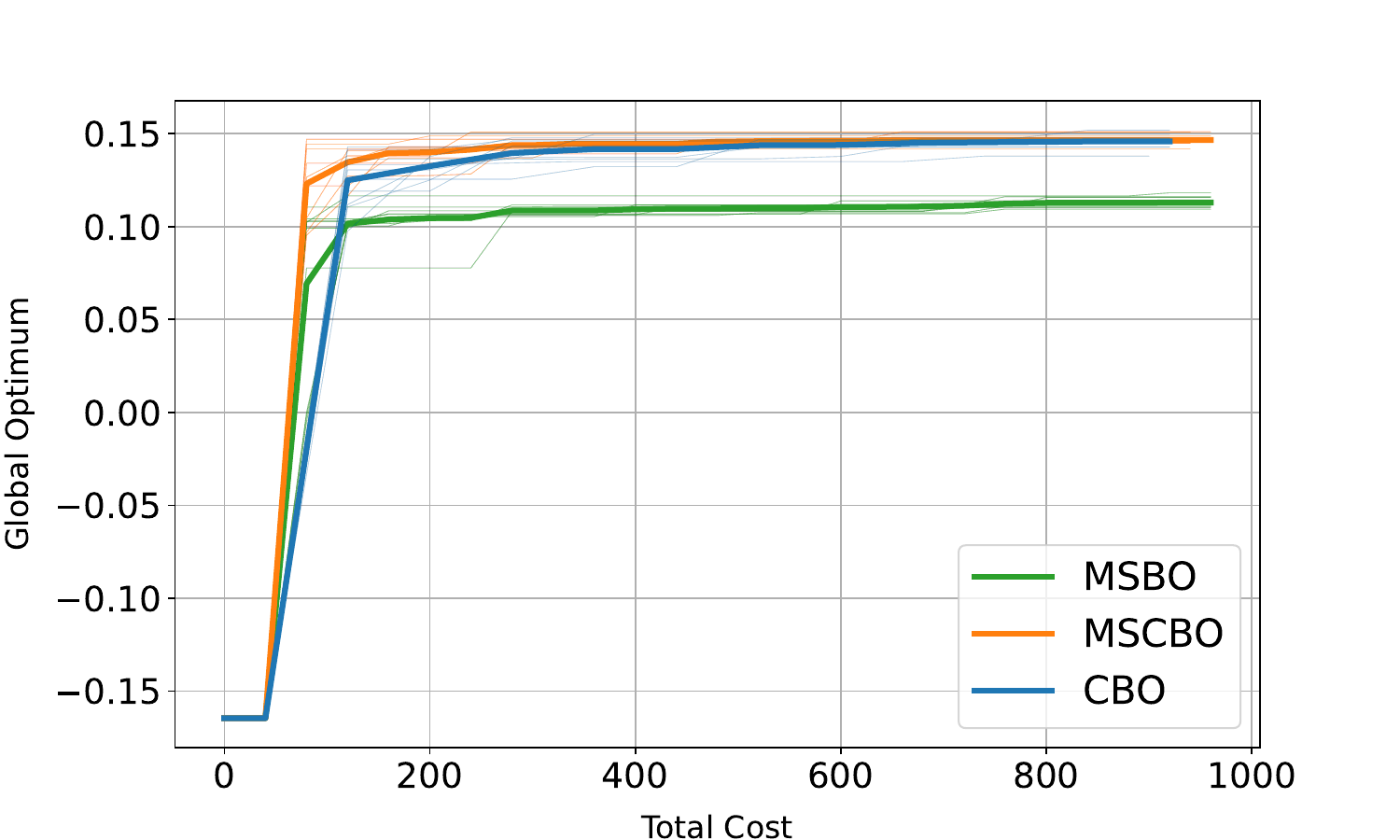}
        \caption{\textbf{Scenario 2}: Added/removed node connections}
    \end{subfigure}
    \hspace{0.05cm}
    \begin{subfigure}{0.45\textwidth}
        \includegraphics[width=\linewidth]{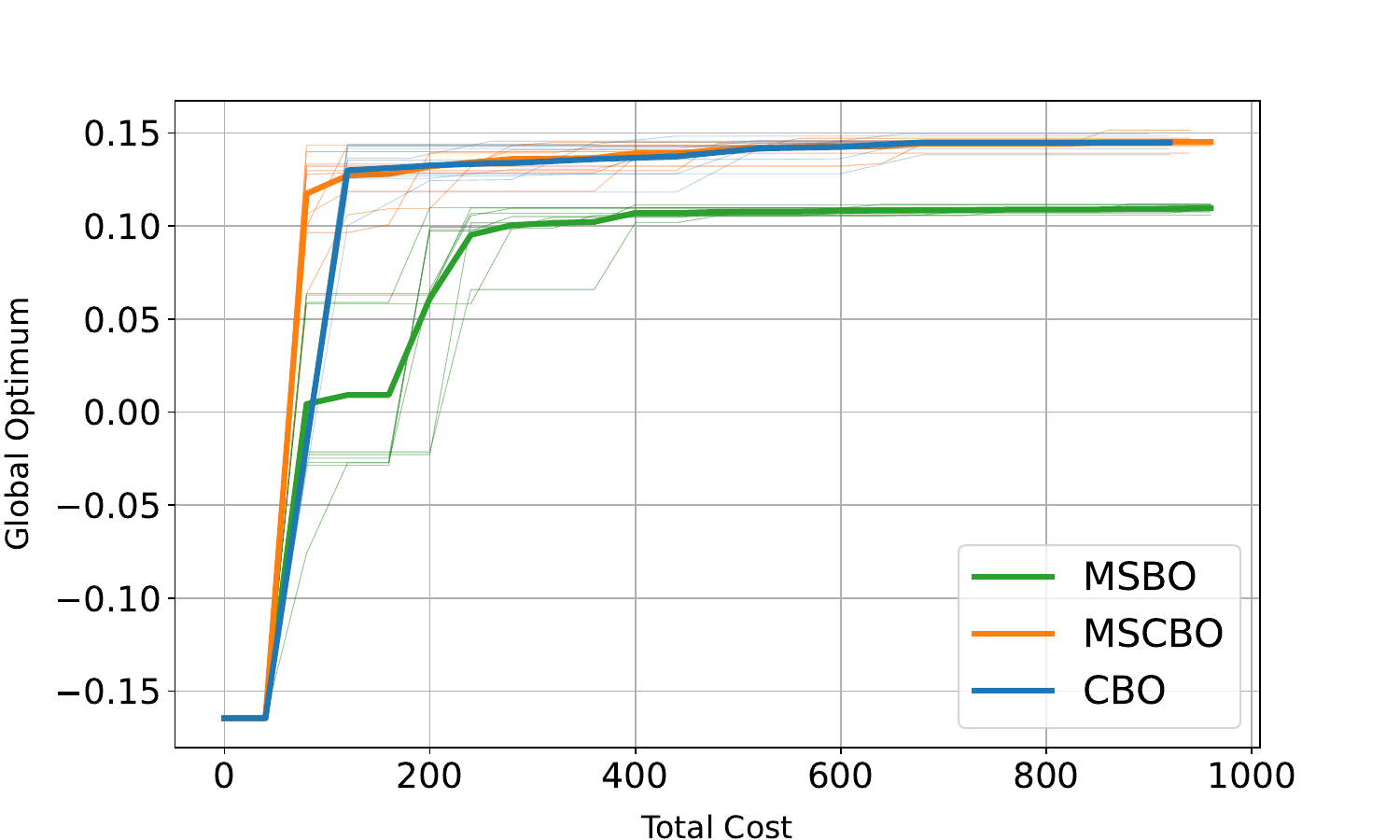}
        \caption{\textbf{Scenario 3}: Added/removed nodes}
    \end{subfigure}

    \caption{Optima over cost for the three different implementations within the Honey Yield graph.}
    \label{fig:beeresults}
\end{figure*}

We treat the network found in the introduced by \citet{smolinski2024causal} in their paper on the effects of various extraneous variables on the honey yield and mortality rate of bee populations. For a more in-depth explanation of the network and variable dynamics, we refer to their paper. As per request of the authors, we do not provide the structural equations of the network variables. \textbf{HY} (honey yield) is denoted as the target variable and we define a optimization problem with the goal of maximizing \textbf{HY} and an exploration set \(\mathcal{V}\) consisting of \(\{\textbf{NO},\textbf{VA}\}\), representing the number of hives merged and the abundance of Varroa (a mite species) in autumn. The example is included as it carries strong ecological validity (Figure \ref{fig:Beehive}). The MIS-optimal set is equal to the exploration set and we do not possess a predefined optimal intervention. The results can be observed from Figure \ref{fig:beeresults}.

We observe both CBO and MSCBO outperforming MSBO in terms of the found optimum and cost-effectiveness, indicating that usage of the causal structure improves performance for this graph. The performance difference between the causal algorithms and standard MSBO appears more pronounced in high-noise situations (scenario 2 and 3). MSCBO generally reaches its optimum at a lower cost than standard CBO, which can be attributed to the halved intervention costs as a result of MSCBO's smart source selection. Thus, the application of MSCBO to examples that use real-world data proves useful, as other traditional methods either lack in cost-effectiveness or convergence quality. 

\subsection{Multi-Armed Bandit (MAB) Example}
We treat the network depicted by \citet{lee2018structural} in their paper on multi-armed bandit interventions. The example network introduced in the paper serves to demonstrate the workings of their POMIS-algorithm and serves as a purely synthetic example. \textbf{Y} is denoted as the target node and we define an optimization problem with the goal of maximizing \textbf{Y} and an exploration set \(\mathcal{V}\) consisting of \(\{\textbf{S},\textbf{W},\textbf{T},\textbf{Z},\textbf{X}\}\). The MIS-optimal set consists of \(\{\textbf{T},\textbf{W}\}\) (with the other minimal candidate set \(\{\textbf{T},\textbf{S}\}\) losing the tiebreaker defined in the Methodology) (Figure \ref{fig:Crop}).  

\begin{figure}[H]
  \centering
  \includegraphics[width= 0.4\textwidth]{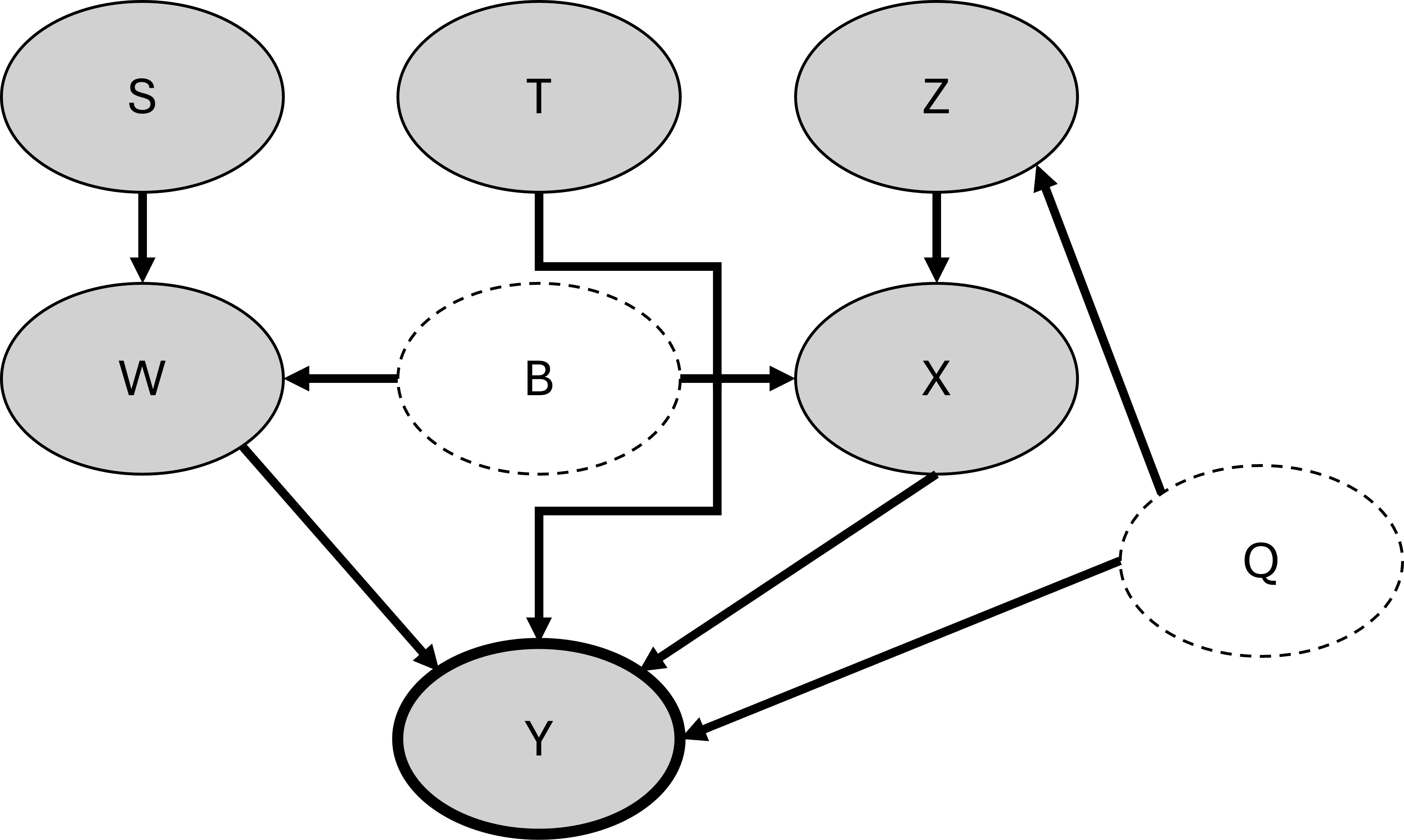}
  \caption{Causal graph depicting a Multi-armed-bandit scenario. Gray nodes represent variables that can be intervened upon, and dashed nodes represent non-manipulative variables. The target variable Y is denoted with a thick-dashed node.}
 \label{fig:Bandit}
\end{figure}

\newpage
Given that the authors of the paper do not provide a structural equation because the paper's focus lies on intervention set pruning (the POMIS-algorithm), we propose a set of simple structural equations.
\[
\begin{aligned}
S &= \mathcal{U}(0, 10) \\
B &= \varepsilon_B \\
Q &= \varepsilon_Q \\
T &= \mathcal{U}(0, 30) \\
W &= 0.6S + 0.4B + \varepsilon_W \\
Z &= 1.5Q + \varepsilon_Z \\
X &= \tanh(0.5Z + 0.3B) + \varepsilon_X \\
Y &= \sigma(0.8W + 0.5T + 1X + 0.6B) + \varepsilon_Y
\end{aligned}
\]

\begin{figure*}[htb]
    \centering
    \begin{subfigure}{0.45\textwidth}
        \includegraphics[width=\linewidth]{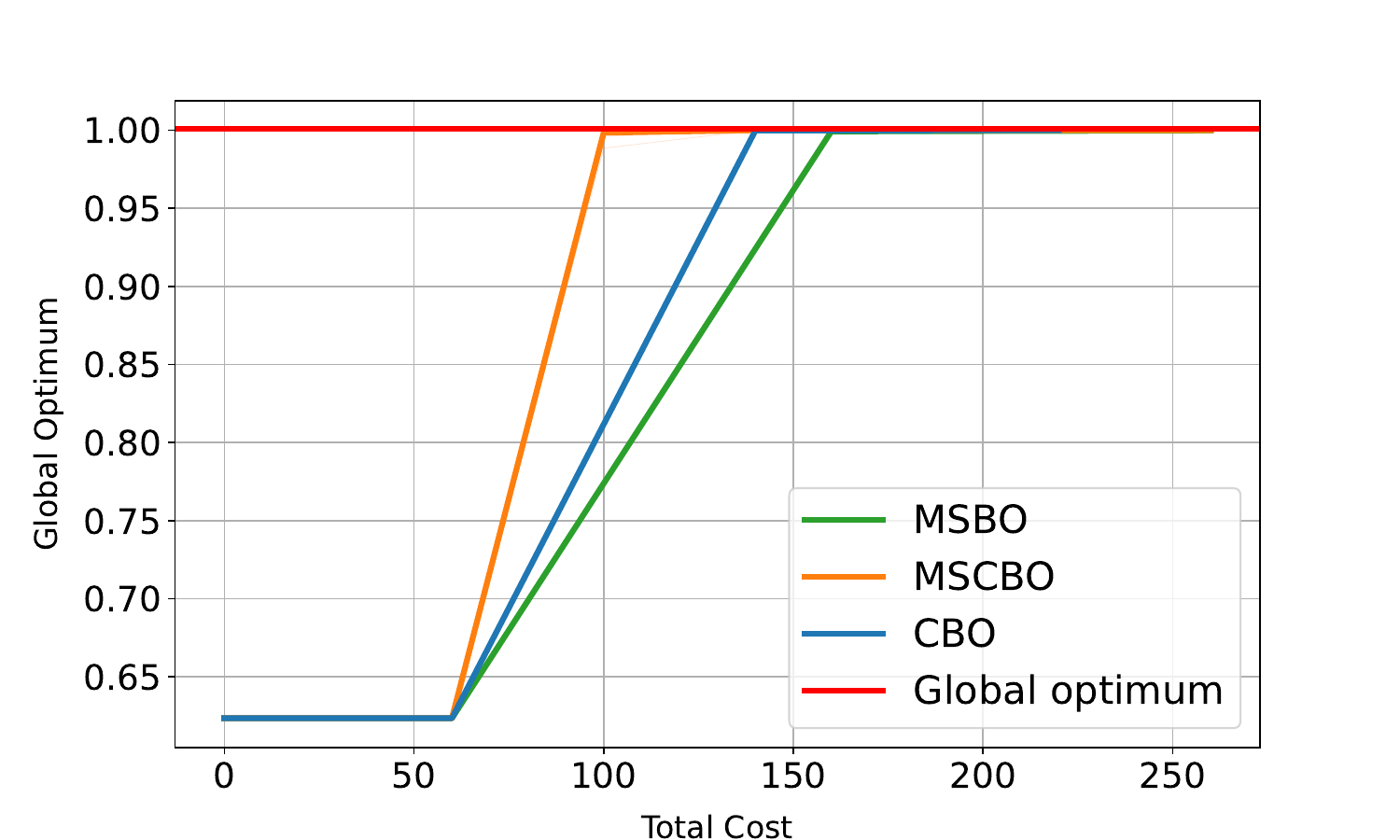}
        \vspace{0.05cm}
        \caption{\textbf{The Base Case}}
    \end{subfigure}
    \hspace{0.05cm}
    \begin{subfigure}{0.45\textwidth}
        \includegraphics[width=\linewidth]{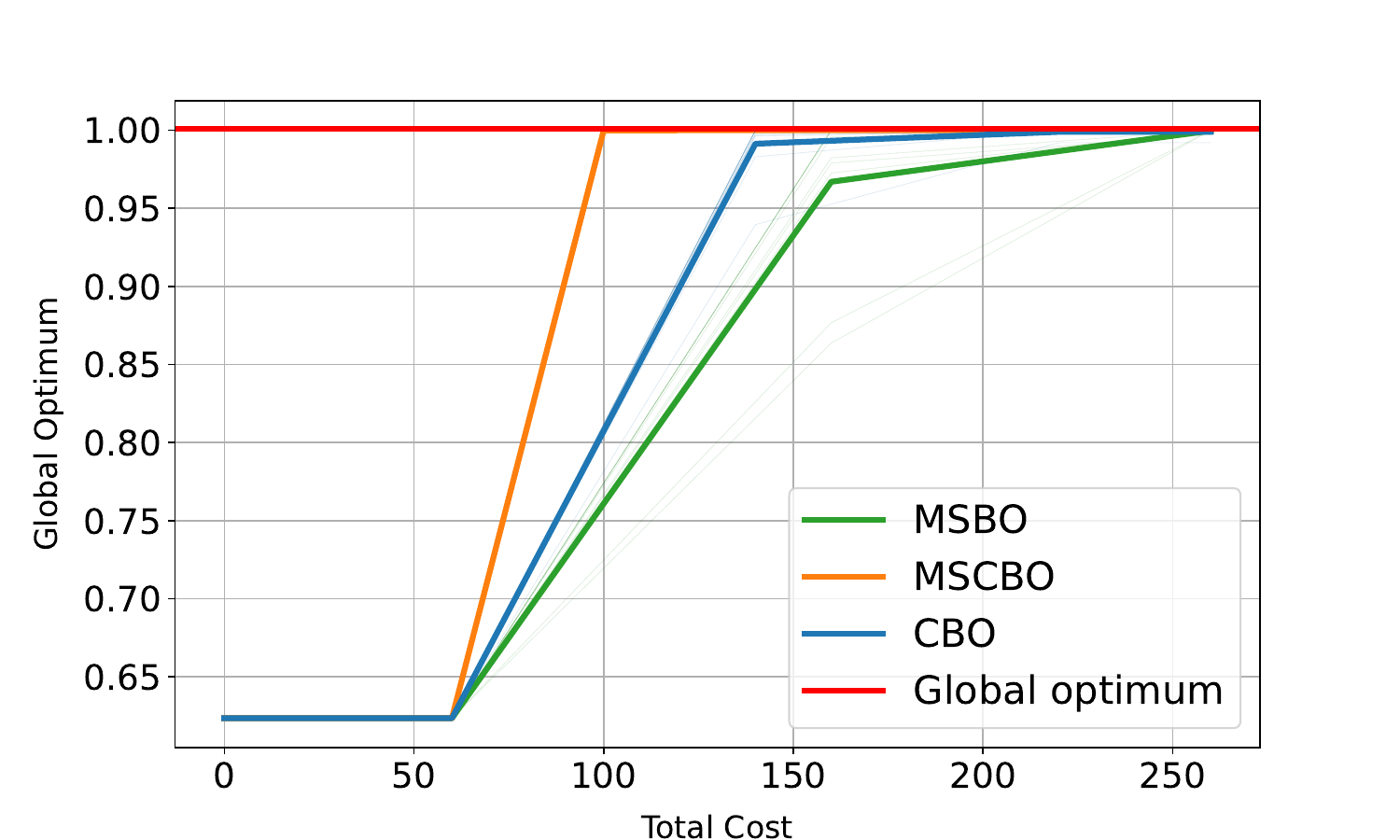}
        \vspace{0.05cm}
        \caption{\textbf{Scenario 1}: altered SCM's}
    \end{subfigure}
    
    \caption{Optima over cost for the three different implementations within the MAB graph.}
    \label{fig:mabresults}
\end{figure*}

The target variable \textbf{Y} is assigned a sigmoid function, as it allows us to inspect whether the global optimum is reached due to the maximum value output of a sigmoid function always being exactly 1. As this network is built for exploration set reduction, it serves as a good sanity check for our algorithm performance. Because MSCBO operates on a smaller intervention set, its cost growth should be comparatively smaller than that of MSBO. The analysis of this example only treats the base case and scenario 1, as altering the connections and nodes within the graph nullifies the concept of the example. The results can be observed from Figure \ref{fig:mabresults}.

All three algorithms consistently reach the optimum in a small number of iterations, which can be attributed to the simplicity of the graph and its mostly linear connections. MSCBO typically reaches the optimal value of 1 before MSBO and CBO as a result of its lower intervention cost, which is exactly half of CBO's (smart source selection) and $\frac{2}{5}$ of MSBO's (exploration set pruning). The example demonstrates the theoretical usefulness of MSCBO in smaller graph structures and the handling of unobserved confounders (nodes \textbf{B} and \textbf{Q}). We observe a slight performance drop-off for CBO and MSBO in scenario 1. MSCBO appears slightly more robust, as it is not affected by the noise.

\newpage
\section{PSA Structural Equations}\label{appendix:PSA_structs}
This section depicts the set of structural equations for the PSA example defined in the main body of the article and sampled from \citet{ferro2015use}.

\[
\begin{aligned}
\text{age} &= \mathcal{U}(55, 75), \\
\text{bmi} &= \mathcal{N}(27.0 - 0.01 \cdot \text{age}, \; 0.7^2), \\
\text{aspirin} &= \sigma(-8.0 + 0.10 \cdot \text{age} + 0.03 \cdot \text{bmi}), \\
\text{statin} &= \sigma(-13.0 + 0.10 \cdot \text{age} + 0.20 \cdot \text{bmi}), \\
\text{cancer} &= \sigma(2.2 - 0.05 \cdot \text{age} + 0.01 \cdot \text{bmi} - 0.04 \cdot \text{statin} + 0.02 \cdot \text{aspirin}), \\
\text{psa} &\sim 
\mathcal{N}\Big(\!6.8 + 0.04 \cdot \text{age} - 0.15 \cdot \text{bmi} - 0.60 \cdot \text{statin} \\
&\quad + 0.55 \cdot \text{aspirin} + 1.00 \cdot \text{cancer}, \; 0.4^2 \!\Big)
\end{aligned}
\]

\section{Extended Theoretical Background}\label{appendix:background}
This section provides a general overview of the causal and multi-source Bayesian optimization (BO) frameworks. Implementation details will be discussed in the methodology. Within the broader class of optimization problems, we focus on Bayesian optimization applied to probabilistic graphical models, specifically Bayesian networks (BNs). These networks are represented as directed acyclic graphs (DAGs), where nodes correspond to random variables, and edges capture conditional dependencies.

In this setting, BO is used to guide interventions on certain variables within the BN to optimize a predefined outcome. Rather than treating the BN as a purely observational model, we assume the ability to manipulate a subset of variables (the exploration set) and analyze the resulting changes in the output node. The goal is to determine the best intervention strategy to maximize or minimize the desired outcome. For instance, in a medical setting, adjusting the dosage of a drug might influence the probability of an inflammatory response. In the absence of optimization techniques, one could intervene on all nodes within the exploration set, so that each combination of node states is attained. By subsequently measuring the effect on the output node, the observer may determine both the min/max value for the output node, as well as the value assignments to the intervention nodes that result in said value. A Gaussian Process (GP) models the objective function \( f(x) \) as:
\[
f(x) \sim \mathcal{GP}(\mu(x), k(x, x')),
\]
where:
\begin{itemize}
    \item \( \mu(x) \): Mean function (commonly assumed \( \mu(x) = 0 \)),
    \item \( k(x, x') \): Covariance (kernel) function, in our case the Radial Basis Function (RBF) kernel:
    \[
    k(x, x') = \sigma^2 \exp\left(-\frac{\|x - x'\|^2}{2\ell^2}\right)
    \].
\end{itemize}

\noindent  Given observations \( D = \{(x_i, y_i)\}_{i=1}^n \), the GP posterior for a new point \( x_* \) is:
\[
\mu_*(x_*) = \mathbf{k}_*^\top (\mathbf{K} + \sigma_n^2 \mathbf{I})^{-1} \mathbf{y},
\]
\[
\sigma^2_*(x_*) = k(x_*, x_*) - \mathbf{k}_*^\top (\mathbf{K} + \sigma_n^2 \mathbf{I})^{-1} \mathbf{k}_*,
\]
where:
\begin{itemize}
    \item \( \mathbf{k}_* = [k(x_*, x_1), \dots, k(x_*, x_n)]^\top \),
    \item \( \mathbf{K} \): Kernel matrix between observed points \( \{x_i\} \),
    \item \( \sigma_n^2 \): Noise variance.
\end{itemize}

\noindent In order to optimize over the GP while balancing exploration and exploitation, an acquisition function \( \alpha(x) \) is introduced:
\[
\alpha(x) = U(\mu_*(x), \sigma_*(x)),
\]
where:
\begin{itemize}
    \item \( \mu_*(x) \): The predicted mean from the surrogate model (e.g., Gaussian Process),
    \item \( \sigma_*(x) \): The predicted standard deviation (uncertainty) from the surrogate model,
    \item \( U \): A utility function that encodes the decision-making strategy.
\end{itemize}

\noindent By maximizing over the acquisition values and performing interventions accordingly, BO can iteratively approach the desired min/max output node value while tracking the corresponding interventions.

\subsection{Causality in Bayesian optimization}
Causal Bayesian optimization (CBO) extends the BO methodology by integrating causality principles and do-calculus with the optimization process. CBO is advantageous over standard BO when optimizing in systems with known causal relationships between variables. By leveraging a causal graph or structural causal model (SCM), CBO focuses on intervenable variables and accounts for how changes propagate through the system, reducing the effective search space and improving efficiency. Unlike BO, which treats all variables as independent, CBO ensures that optimization respects the underlying causal structure. A Structural Causal Model (SCM) is defined as a tuple \( \mathcal{M} = (\mathcal{V}, \mathcal{U}, \mathcal{F}, P(\mathcal{U})) \), where:
\begin{itemize}
    \item \( \mathcal{V} \): Set of observed variables, such as \( X_1, X_2, \dots, X_d, Y \).
    \item \( \mathcal{U} \): Set of unobserved (latent) variables.
    \item \( \mathcal{F} \): Set of structural equations \( V_i = f_i(\text{Pa}(V_i), U_i) \), where \( \text{Pa}(V_i) \subseteq \mathcal{V} \) are the parents of \( V_i \) in the causal graph.
    \item \( P(\mathcal{U}) \): Joint distribution over the unobserved variables.
\end{itemize}

\noindent Given an SCM, the goal in Causal Bayesian Optimization (CBO) is to maximize the effect of an \textit{intervention} \( \text{do}(X=x) \) on the target variable \( Y \). Thus, the optimization problem in CBO is defined as:
\[
x^* = \arg\max_{x \in \mathcal{X}} \mathbb{E}[Y \mid \text{do}(X = x)],
\]
where:
\begin{itemize}
    \item \( X \): Set of intervenable variables, a subset of \( \mathcal{V} \),
    \item \( \text{do}(X = x) \): Represents an intervention that sets \( X \) to \( x \), removing the causal influence of \( \text{Pa}(X) \) on \( X \),
    \item \( \mathbb{E}[Y \mid \text{do}(X = x)] \): The expected value of \( Y \) given the intervention \( \text{do}(X = x) \), derived using the interventional distribution: 
    \begin{align}
        P(Y \mid \text{do}(X = x)) &= \int P(Y \mid X = x, \text{Pa}(Y)) \notag \\
        &\quad \cdot P(\text{Pa}(Y) \mid \text{do}(X = x)) \, d\text{Pa}(Y).
        \notag
    \end{align}
\end{itemize}

\noindent Compared to the traditional BO approach, CBO makes use of a Causal Gaussian Process (GP) model, and an \( \epsilon \)-greedy policy. To optimize the objective, the surrogate Causal Gaussian Process is used to approximate \( \mathbb{E}[Y \mid \text{do}(X = x)] \), and is trained using interventions.

The acquisition function in CBO, \( \alpha(x) \), guides the choice of the next intervention:
\[
x_{\text{next}} = \arg\max_{x \in \mathcal{X}} \alpha(x),
\]
where \( \alpha(x) \) balances exploration (uncertainty about \( \mathbb{E}[Y \mid \text{do}(X = x)] \)) and exploitation (maximizing the estimated \( \mathbb{E}[Y \mid \text{do}(X = x)] \)).

\subsection{Multi-source optimization}

Multi-source Bayesian optimization (MSBO) is a further extension of the BO framework which encompasses optimization over multiple (possibly noisy) information sources, as opposed to the single-source optimization used in both CBO and BO. In MSBO, the objective is to optimize a continuous function \( g: \mathcal{D} \to \mathbb{R} \), where \(\mathcal{D}\) is the design space. Thus, \(\mathcal{D}\) encompasses all possible value assignments \( \text{do}(X = x) \) to the intervention sets \( X \) over all information sources. The goal is to identify the optimal design \( x^* \) such that:
\[
x^* = \arg \max_{x \in \mathcal{D}} g(x).
\]

\noindent To achieve this, \( M \) information sources, denoted as \( IS_1, IS_2, \dots, IS_M \), are leveraged. These sources provide observations about \( g \), but they may be biased and/or noisy. Each source \( IS_\ell(x) \) provides independent observations modeled as:
\[
IS_\ell(x) \sim \mathcal{N}(f_\ell(x), \lambda_\ell(x)),
\]
where \( f_\ell(x) \) represents the mean prediction from the source, and \(\lambda_\ell(x)\) is the variance of the observations. These sources are often referred to as \textit{surrogates} or \textit{auxiliary tasks}, as they approximate the true objective \( g(x) \), which is the primary task. The relationship between the true function \( g(x) \) and the approximations provided by the sources is captured by the discrepancy term:
\[
\delta_\ell(x) = g(x) - f_\ell(x).
\]
We assume that this discrepancy follows a Gaussian process (GP) prior:
\[
\delta_\ell(x) \sim \mathcal{GP}(\mu_\ell, \Sigma),
\]
where \(\mu_\ell(x)\) is the prior mean, which is often set to \( 0 \) unless domain-specific knowledge suggests otherwise. This results in a model for each source:
\[
f_\ell(x) = f_0(x) + \delta_\ell(x),
\]
where \( f_0(x) \sim \mathcal{GP}(\mu_0, \Sigma_0) \) is a GP prior for the true function \( g(x) \). The covariance structure between sources is defined as:
    
\begin{align}
\Sigma((\ell, x), (m, x')) &= \text{Cov}(f_\ell(x), f_m(x')) \notag \\ &= \Sigma_0(x, x') + \delta_{\ell m} \cdot \tau_\ell^2(x, x'),
\notag
\end{align}

\noindent where \(\delta_{\ell m}\) is the Kronecker delta, which is \( 1 \) if \(\ell = m\) and \( 0 \) otherwise, and \(\tau_\ell^2(x, x')\) captures variance specific to source \(\ell\).

The variance of observations from each source, \(\lambda_\ell(x)\), and the cost of querying a source, \( c_\ell(x) \), are assumed to be known and continuously differentiable. These costs may be provided by domain experts or estimated alongside the model parameters. The cost function \( c_\ell(x) \) enables the framework to balance the cost of querying a source with the potential information gain, which is particularly important in settings where interventions are costly or time-consuming. By incorporating multiple information sources, this framework allows for efficient exploration of the design space \(\mathcal{D}\), even when direct evaluations of \( g(x) \) are expensive or simply impossible. The probabilistic structure enables principled decision-making by accounting for both uncertainty and cost across sources, thereby guiding the search for the optimal design \( x^* \).

\section{E. coli network structural overview}\label{appendix:ecoligraphoverview}
\begin{figure*}[htb]
  \centering
  \includegraphics[width=\textwidth]{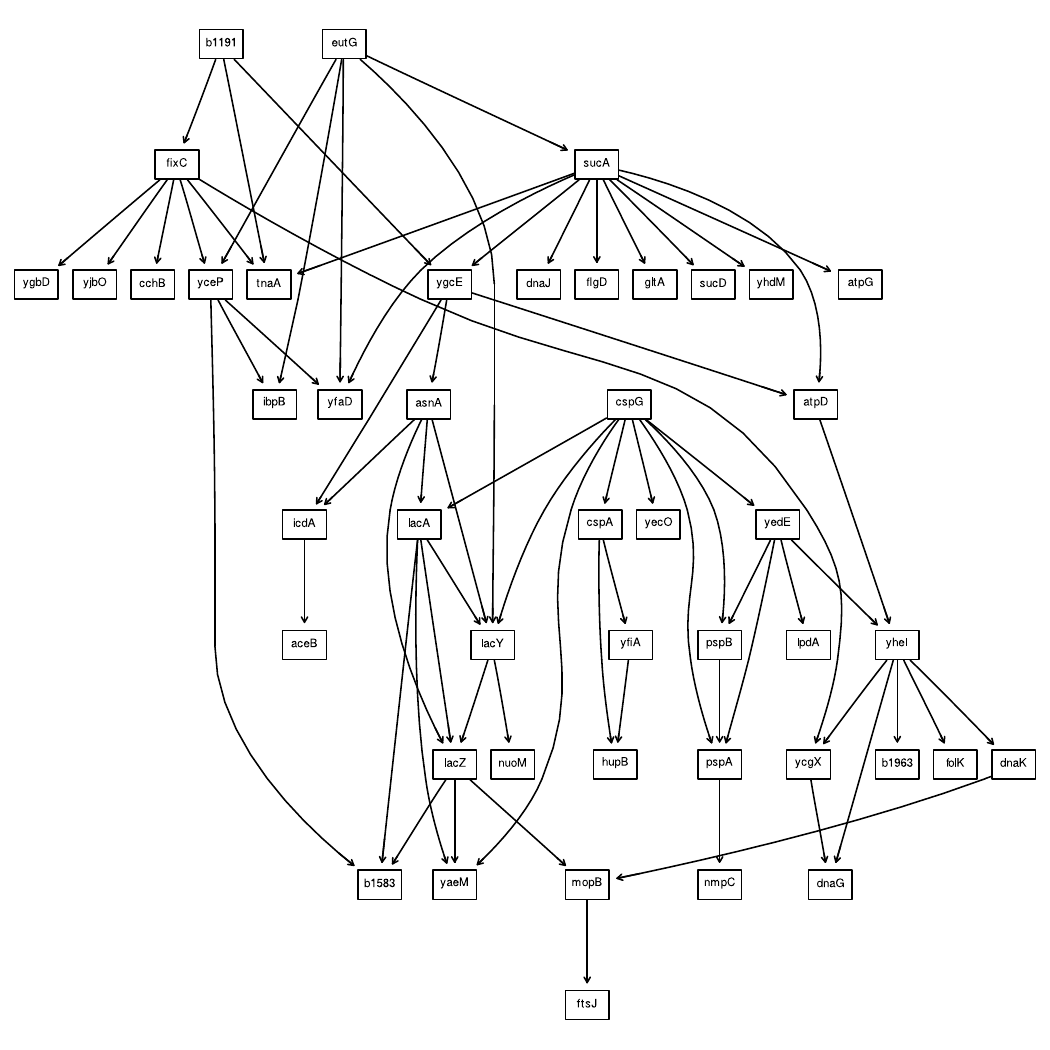}
  \caption{Causal graph depicting the E. coli scenario.}
 \label{fig:Ecolioverview}
\end{figure*}

This section contains a structural overview of the E. coli network as found in the Bayesian Network Repository \cite{bnrepository} (Figure \ref{fig:Ecolioverview}).

\end{document}